\documentclass{article}

\usepackage[utf8]{inputenc} 
\usepackage[T1]{fontenc}    
\usepackage{hyperref}       
\usepackage{url}            
\usepackage{booktabs}       
\usepackage{amsfonts}       
\usepackage{amsmath}
\usepackage{subcaption}

\usepackage{nicefrac}       
\usepackage{microtype}      
\usepackage{amsmath}
\usepackage{amssymb}
\usepackage{helvet}
\usepackage{courier}
\usepackage{graphicx}
\usepackage{booktabs}
\usepackage{xcolor}
\usepackage{float}
\usepackage{multirow}
\usepackage{comment,amsmath,amssymb,color,float,enumitem,overpic,authblk,abstract,cite,geometry,xifthen,tabularx,fancyhdr}
\usepackage{setspace}
\date{}

\title{Human detection of machine manipulated media} 
\author[1]{Matthew Groh}
\author[1]{Ziv Epstein}
\author[1,2]{Nick Obradovich}
\author[1,2]{Manuel Cebrian$^*$}
\author[1,2]{Iyad Rahwan\footnote{To whom correspondence should be addressed: \emph{cebrian@mit.edu, irahwan@mit.edu}}}
\affil[1]{Media Laboratory, Massachusetts Institute of Technology, Cambridge, Massachusetts, USA }
\affil[2]{Center for Humans \& Machines, Max Planck Institute for Human Development, Berlin, Germany}
\begin{document}
\maketitle
\thispagestyle{fancy}
\begin{abstract}
\begin{quote}

Recent advances in neural networks for content generation  enable artificial intelligence (AI) models to generate high-quality media manipulations. Here we report on a randomized experiment designed to study the effect of exposure to media manipulations on over 15,000 individuals' ability to discern machine-manipulated media. We engineer a neural network to plausibly and automatically remove objects from images, and we deploy this neural network online  with a randomized experiment where participants can guess which image out of a pair of images has been manipulated. The system provides participants feedback on the accuracy of each guess. In the experiment, we randomize the order in which images are presented, allowing causal identification of the learning curve surrounding participants' ability to detect fake content. We find sizable and robust evidence that individuals learn to detect fake content through exposure to manipulated media when provided iterative feedback on their detection attempts. Over a succession of only ten images, participants increase their rating accuracy by over ten percentage points. We then investigate factors that potentially moderate rates of learning and find that image quality, proportion of image modified, and user device type -- among other factors -- may all play important roles in learning to detect manipulated media. Our study provides initial evidence that human ability to detect fake, machine-generated content may increase alongside the prevalence of such media online.
\end{quote}
\end{abstract}
\section*{Introduction}
The recent emergence of artificial intelligence (AI) powered media manipulations has widespread societal implications for journalism and democracy\cite{chesney2018deep}, national security\cite{allen2017artificial}, and art \cite{hertzmann2018can}. AI models have the potential to scale misinformation to unprecedented levels by creating various forms of synthetic media \cite{lazer2018science}. For example, AI systems can synthesize realistic video portraits of an individual with full control of facial expressions including eye and lip movement\cite{thies2016face2face, suwajanakorn2017synthesizing, kim2018deep, DBLP:journals/corr/SaitoWHNL16, garrido2015vdub}, can clone a speaker's voice with few training samples and generate new natural sounding audio of something the speaker never previously said\cite{arik2018neural}, can synthesize visually indicated sound effects \cite{DBLP:journals/corr/OwensIMTAF15}, can generate high quality, relevant text based on an initial prompt \cite{gpt2}, can produce photorealistic images of a variety of objects from text inputs\cite{DBLP:journals/corr/NguyenYBDC16,brock2018large,karras2018style}, and can generate photorealistic videos of people expressing emotions from only a single image \cite{averbuch2017bringing, Zakharov2019FewShotAL}. The technologies for producing entirely machine-generated, fake media online are rapidly outpacing the ability to manually detect and respond to such media.

\begin{figure}[H]
    \centering
    \includegraphics[width=0.98\textwidth]{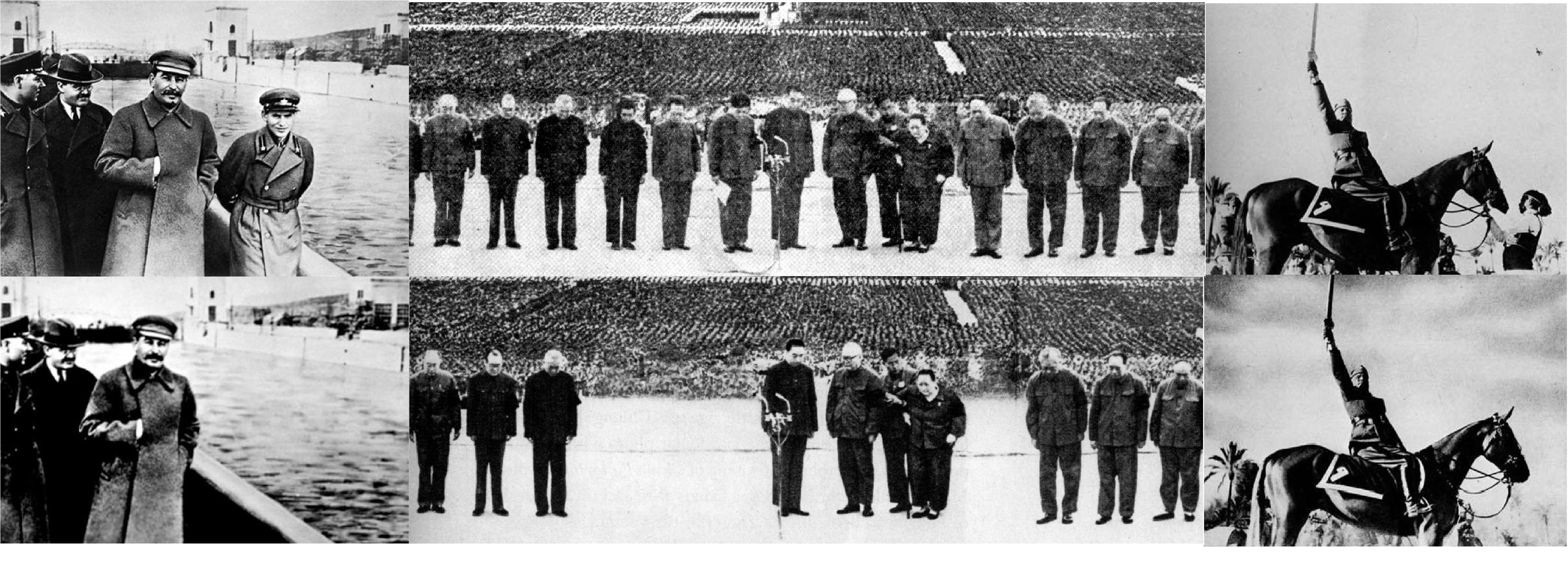}
    \caption{Photographic manipulation has long been a tool of fascist governments. On the left, Joseph Stalin is standing next to Nikolai Yezhov who Stalin later ordered to be executed and disappeared from the photograph. In the middle, Mao Zedong is standing beside the ``Gang of Four'' who were arrested a month after Mao's death and subsequently erased. On the right, Benito Mussolini strikes a heroic pose on a horse while his trainer holds the horse steady.}
    \label{fig:leadersxx}
\end{figure}

Media manipulation and misinformation are topics of considerable interest within the computational and social sciences \cite{Vosoughi1146, benkler2018network, cooke2017posttruth, marwick2017media}, partially because of their historical significance. For a particular kind of media manipulation, there's a Latin term, \textit{damnatio memoriae}, which refers to the erasure of an individual from official accounts, often in service of dominant political agendas. The earliest known instances of \textit{damnatio memoriae} were discovered in ancient Egyptian artifacts and similar patterns of removal have appeared since \cite{varner2004monumenta, freedberg1989power}. Figure SI\ref{fig:leadersxx} presents iconic examples of \textit{damnatio memoriae} throughout modern history. Historically, visual and audio manipulations required both skilled experts and a significant investment of time and resources. Today, an AI model can produce photorealistic manipulations nearly instantaneously, which magnifies the potential scale of misinformation. This growing capability
calls for understanding individuals' abilities to differentiate between real and fake content. 

\begin{figure*}[h!]

\includegraphics[width=0.24\linewidth]{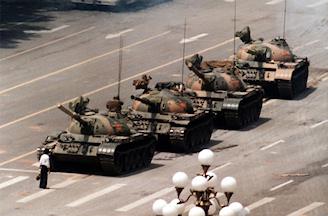}
\includegraphics[width=0.24\linewidth]{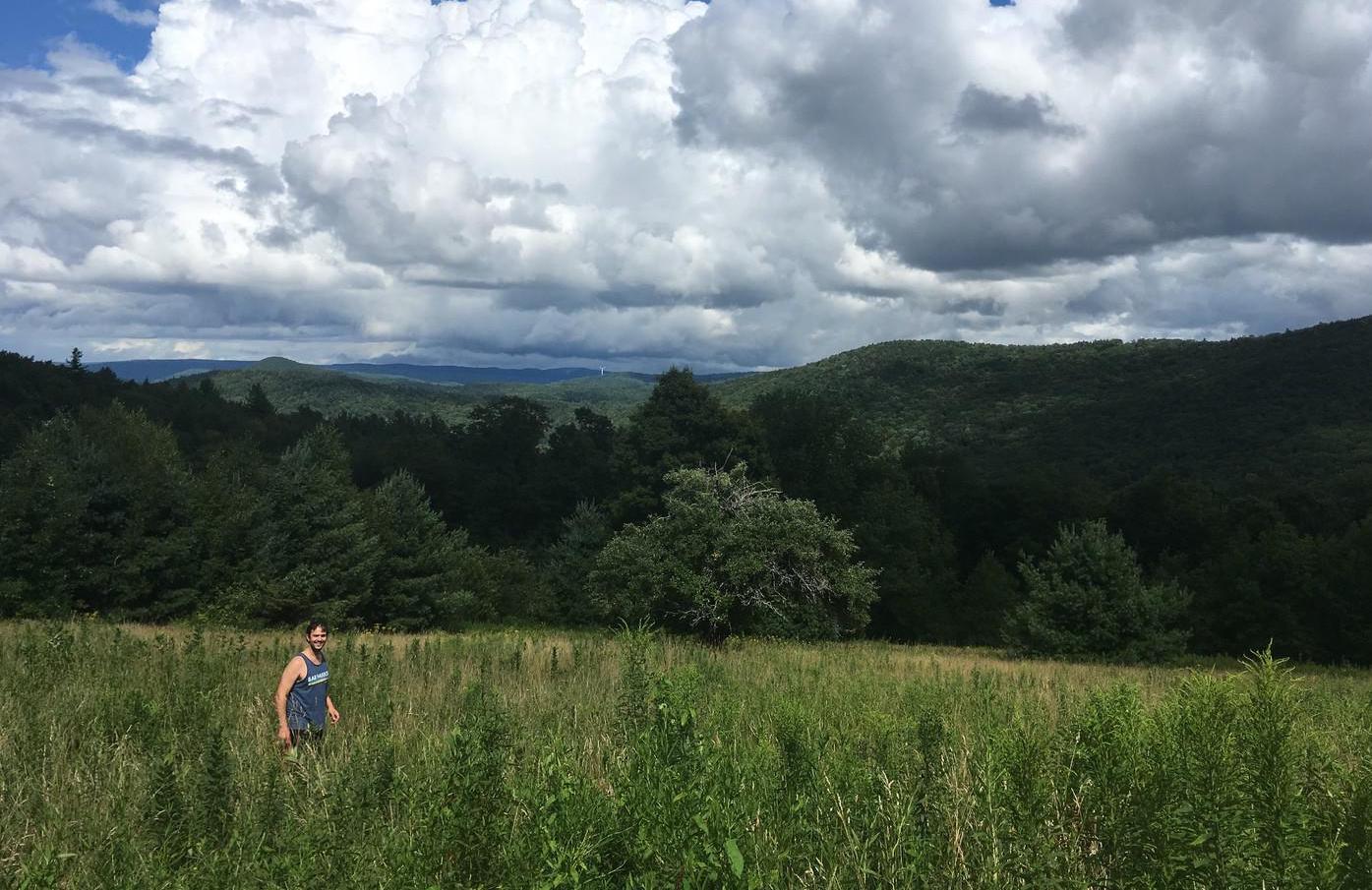}
\includegraphics[width=0.24\linewidth]{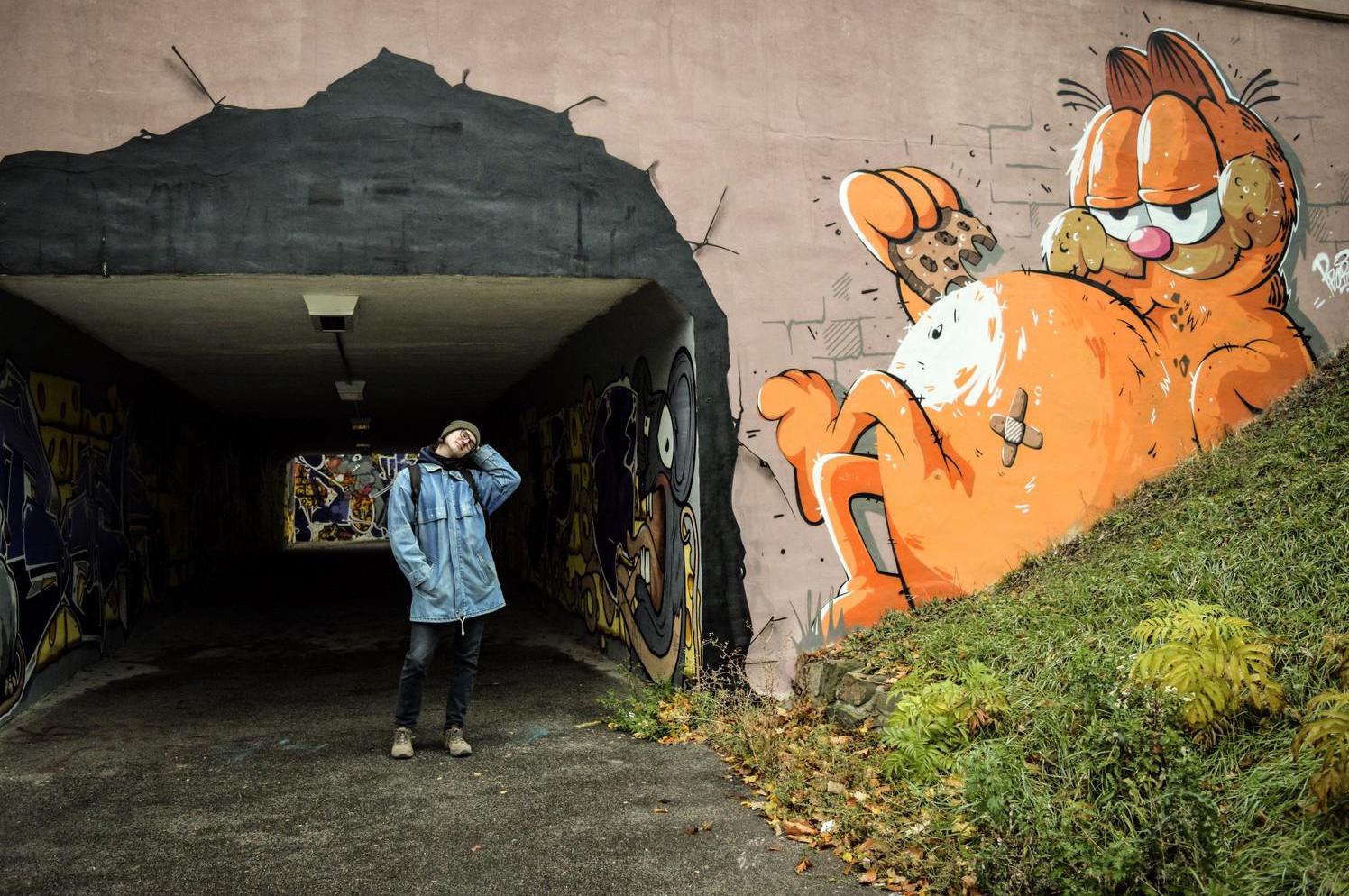}
\includegraphics[width=0.24\linewidth]{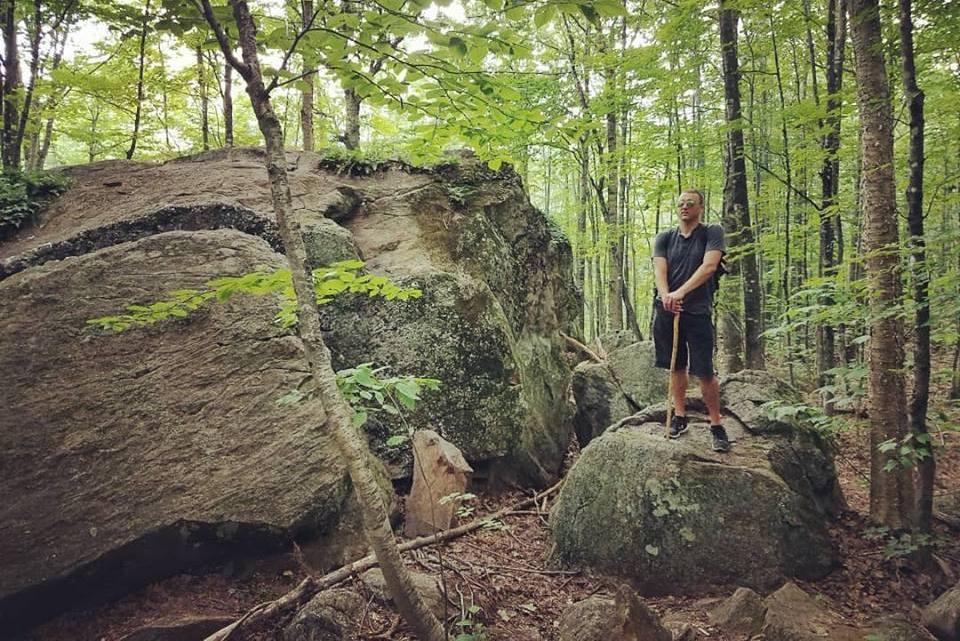}

\includegraphics[width=0.24\linewidth]{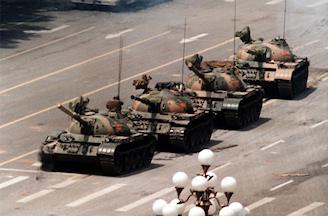}
\includegraphics[width=0.24\linewidth]{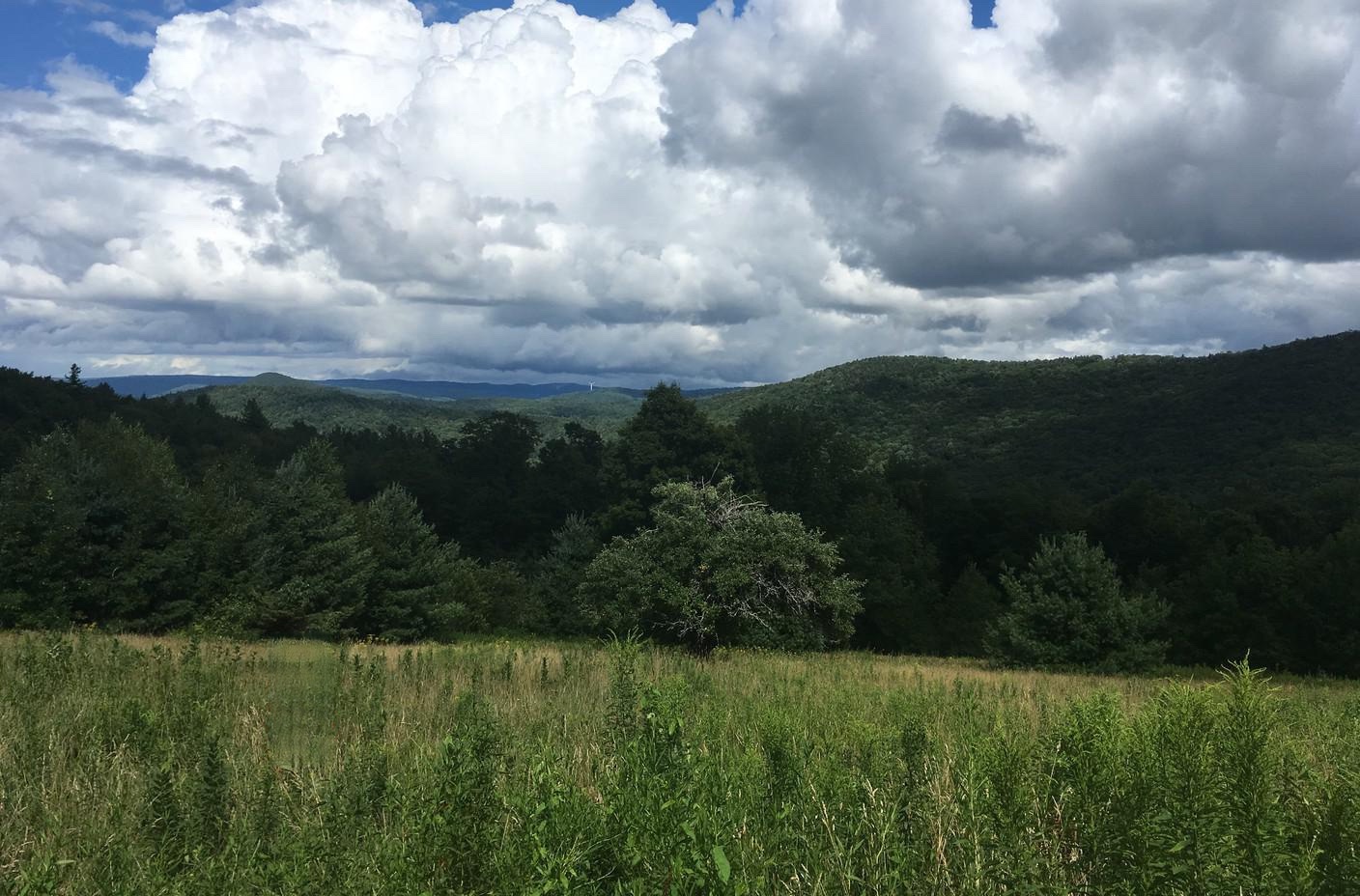}
\includegraphics[width=0.24\linewidth]{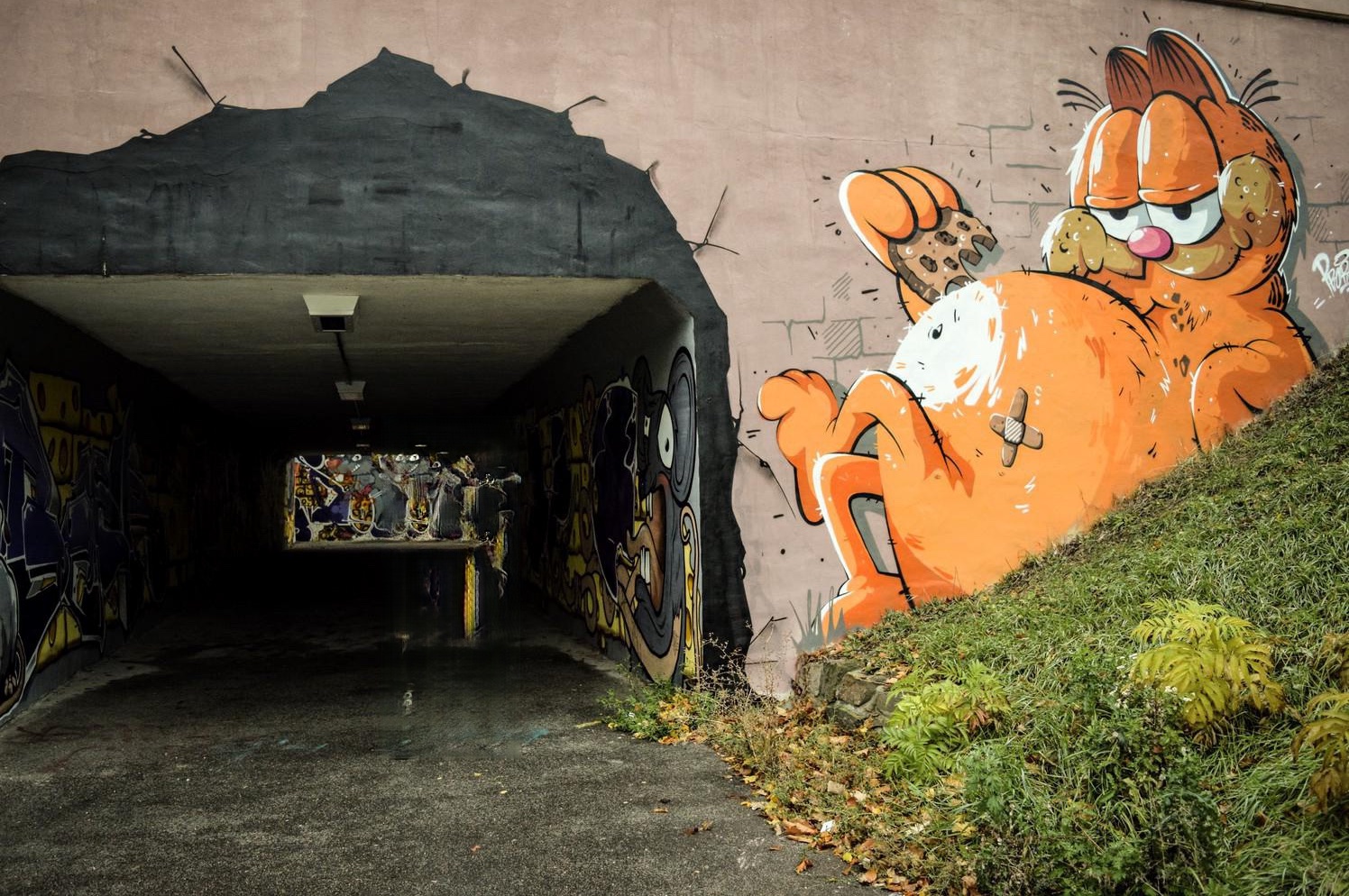}
\includegraphics[width=0.24\linewidth]{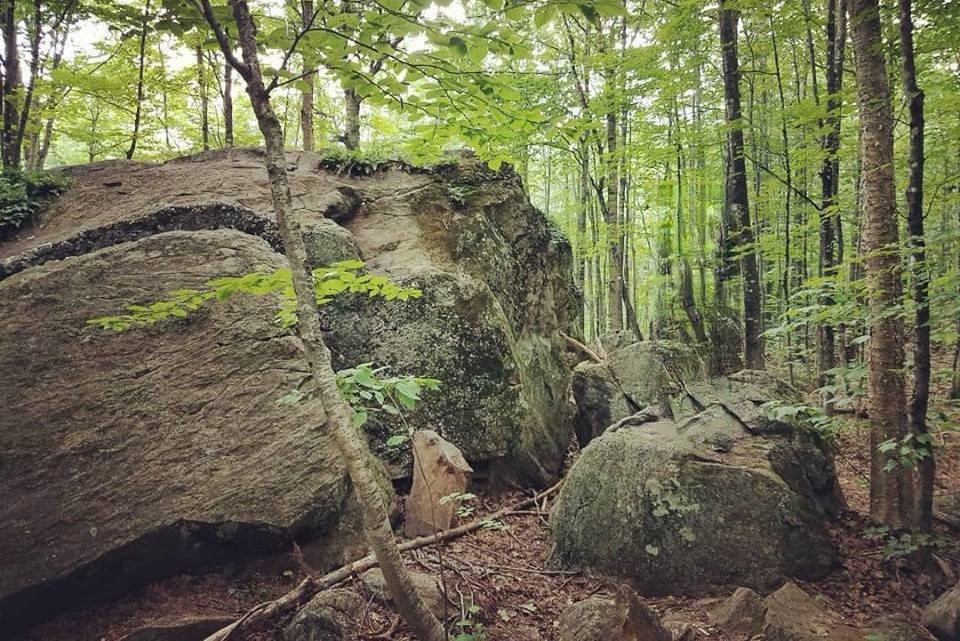}

\caption{Examples of original images on the top row and manipulated images on the bottom row.}
\label{fig:imagesttt}%
\end{figure*}

To interrogate these questions directly, we engineer an AI system for photorealistic image manipulation and host the model and its outputs online as an experiment to study participants' abilities to differentiate between unmodified and manipulated images. Our AI system consists of an end-to-end neural network architecture that can plausibly disappear objects from images. For example, consider an image of a boat sailing on the ocean. The AI model detects the boat, removes the boat, and replaces the boat's pixels with pixels that approximate what the ocean might have looked like without the boat present. Figure \ref{fig:imagesttt} presents four examples of participant submitted images and their transformations. We host this AI model and its image outputs on a custom-designed website called Deep Angel. Since Deep Angel launched in August 2018, over 110,000 individuals have visited the website and interacted with the model and its outputs. Within the Deep Angel platform, we embedded a randomized experiment to examine how repeated exposure to machine-manipulated images affects individuals' ability to accurately identify manipulated imagery. 


\section*{Experimental Design}

\subsection*{User Interface}

In the ``Detect Fakes'' feature on Deep Angel, individuals are presented with two images and asked a single question: ``Which image has something removed by Deep Angel?'' See Figure~\ref{fig:uidetect} in the Supplementary Information for a screenshot of this interaction. One image has an object removed by our AI model. The other image is an unaltered image from the 2014 MS-COCO data \cite{lin2014microsoft}. After a participant answers the question by selecting an image, the manipulated image is revealed to the participant and the participant is offered the option to try again on a new pair of images. 


\subsection*{Usage}

Most participants interacted with ``Detect Fakes'' multiple times; the interquartile range of the number of guesses per participant is from 3 to 18 with a median of 8. Each interaction followed the same randomization with replacement, which ensured that the images displayed did not depend on what the individual had previously seen. 

\begin{figure*}[t!]
\centering
\begin{subfigure}[t]{0.49\textwidth}
    \centering
    \includegraphics[width=0.99\textwidth]{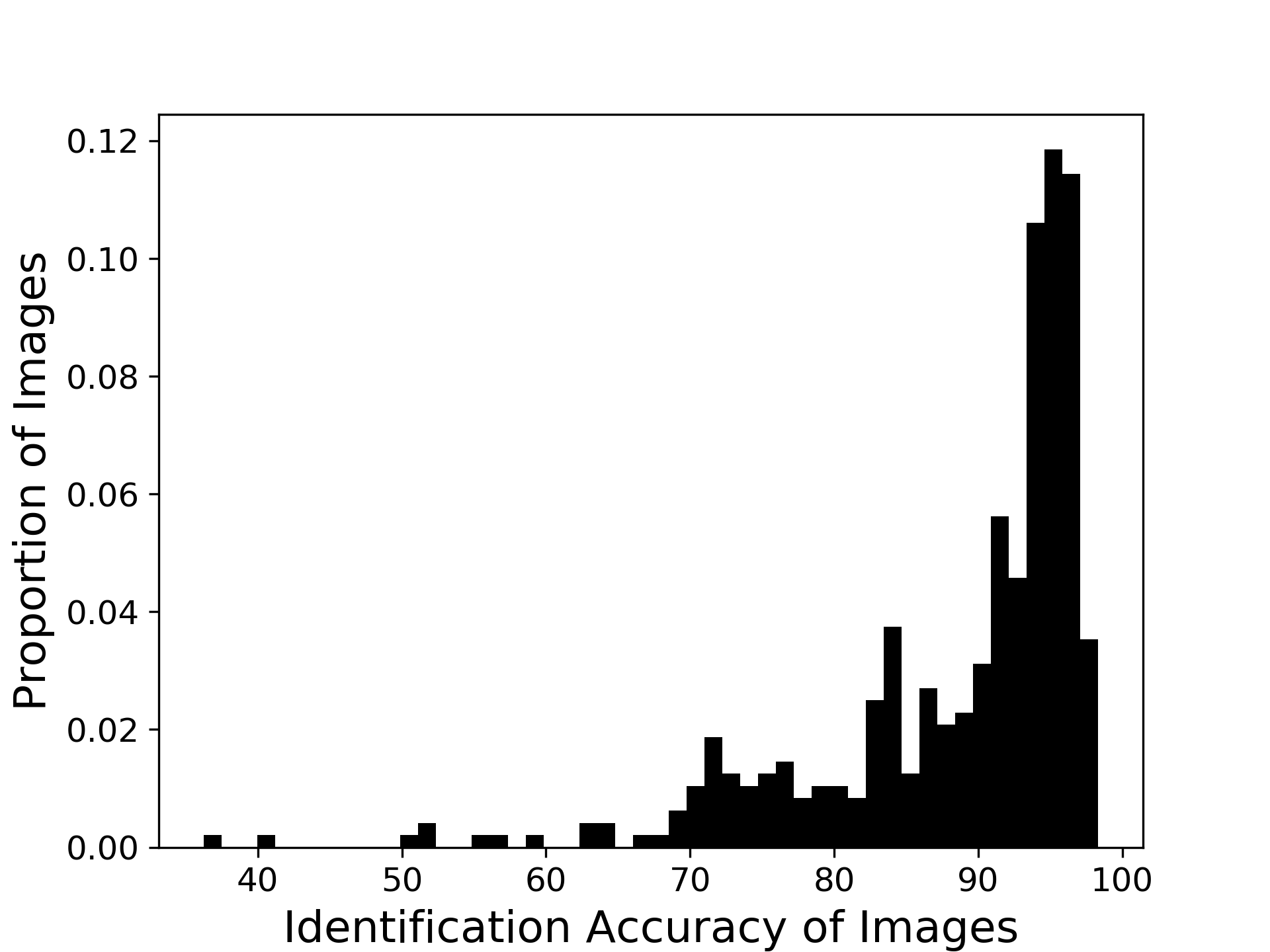}
    \caption{}
\end{subfigure}%
~
\begin{subfigure}[t]{0.49\textwidth}
    \centering
    \includegraphics[width=0.99\textwidth]{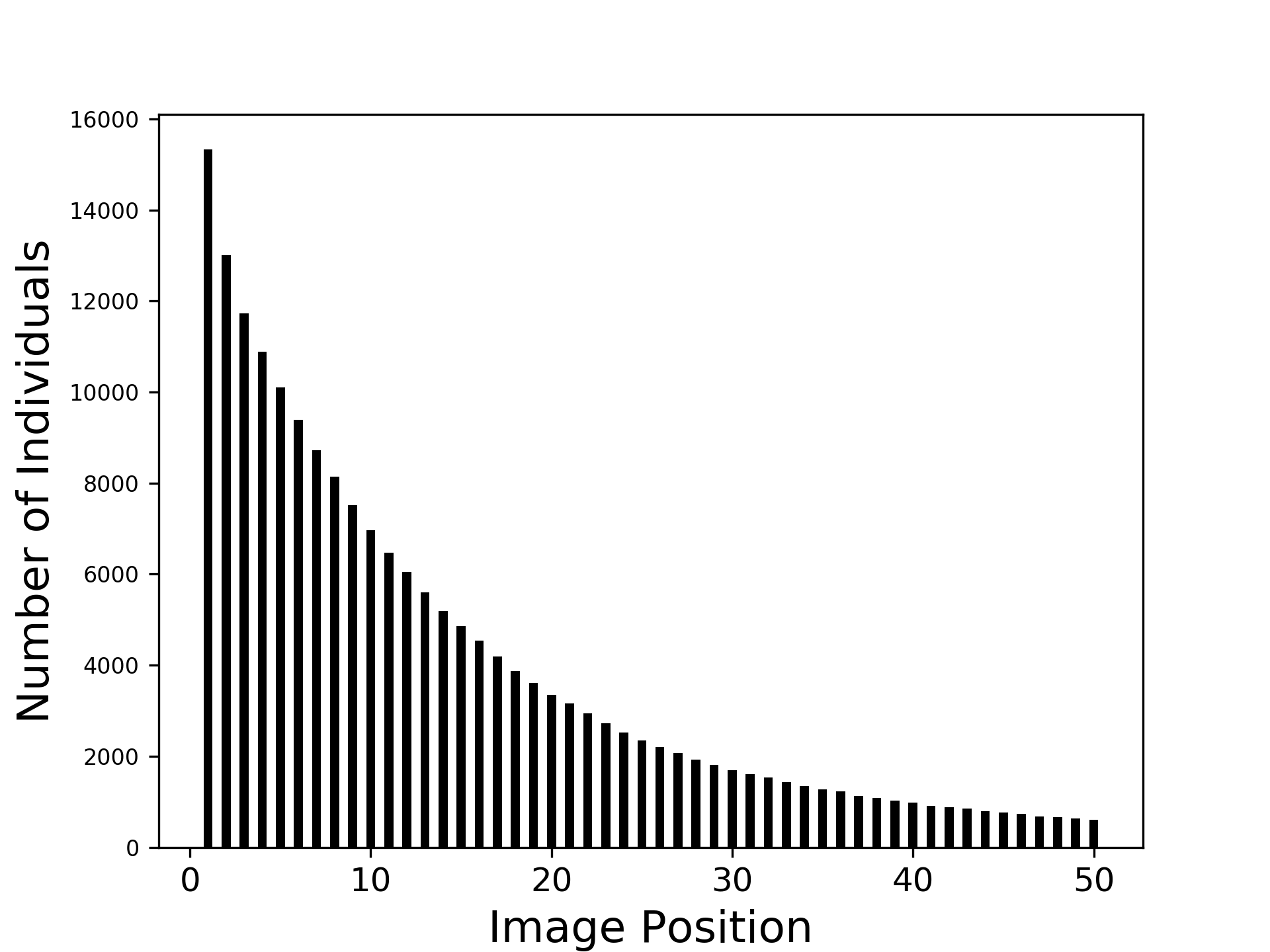}
    \caption{}
\end{subfigure}%
\caption{(a) Histogram of mean identification accuracies by participants per image (b) Bar chart plotting number of individuals over image position.}
\label{fig:statsiguess}%
\end{figure*}

From August 2018 to May 2019, 242,216 guesses were submitted from 16,542 unique IP addresses with a mean identification accuracy of 86\%. Deep Angel did not require participant sign-in, so we study participant behavior under the assumption that each IP address represents a single individual. 7,576 participants submitted at least 10 guesses. Each image appears as the first image an average of 35 times and the tenth image an average of 15 times. In the sample of participants who saw at least ten images, the mean percentage correct classification is 78\% on the first image seen and 88\% on the tenth image seen. The majority of manipulated images were identified correctly more than 90\% of the time. Figure \ref{fig:statsiguess}a shows the distribution of identification accuracy over images, and Figure \ref{fig:statsiguess}b shows the distribution of image positions seen over participants.

By plotting participant identification accuracy against the order in which participants see images, Figure \ref{fig:regression_plot}a reveals a logarithmic relationship between accuracy and overall exposure to manipulated images. Accuracy increases fairly linearly over the first ten images after which accuracy plateaus around 88\%.

\subsection*{Randomization}

We randomly select the ``Detect Fakes'' images from two samples of images. One sample contains 440 images manipulated by Deep Angel that participants submitted to be shared publicly. The other pool of images contains 5,008 images from the MS-COCO dataset \cite{lin2014microsoft}. 
Such randomization at the image dyad level is equivalent to randomization of the image position - the order in which images appear to the participant. Based on the randomized image position, we can causally evaluate the effect of image position on rating accuracy. We test the causal effects with the following linear probability models:



\begin{equation}
y_{i,j} = \alpha X_{i,j} + \beta \log(T_{i_n}) + \mu_i + \nu_j + \epsilon_{i,j}
\label{equation:one}
\end{equation}

and

\begin{equation}
y_{i,j} = \alpha X_{i,j} + 
\beta_1 T_{i_1}  +
\beta_2 T_{i_2}  +
\beta_3 T_{i_3}  +
... + 
\beta_9 T_{i_9}  +
\beta_{10} T_{i_{10}}  +
\mu_i + \nu_j + \epsilon_{i,j}
\label{equation:two}
\end{equation}

where $y_{i,j}$ is the binary accuracy (correct or incorrect guess) of participant $j$ on manipulated image $i$. $X_{i,j}$ represents a matrix of covariates, $T_{i_n}$ represents the order $n$ in which manipulated image $i$ appears to participant $j$,
$\mu_i$ represents the manipulated image fixed effects, $\nu_j$ represents the participant fixed effects, and $\epsilon_{i,j}$ represents the error term. 
The first model fits a logarithmic transformation of $T_{i_n}$ to $y_{i,j}$. The second model estimates treatment effects separately for each image position. Both models use Huber-White (robust) standard errors, and errors are clustered at the image level.
\section*{Results}

With 242,216 observations, we run an ordinary least squares regression with user and image fixed effects on the likelihood of guessing the manipulated image correctly. The results of these regressions are presented in Tables \ref{table:ols_1} and \ref{table:ols_2} in the Appendix. 
Each column in Table \ref{table:ols_1} and \ref{table:ols_2} adds an incremental filter to offer a series of robustness checks. The first column shows all observations. The second column drops all users who submitted fewer than 10 guesses and removes all control images where nothing was removed. The third column drops all observations where a user has already seen a particular image. The fourth column drops all images qualitatively judged as below very high quality. 

\begin{figure*}[t!]
\centering
\begin{subfigure}[t]{0.49\textwidth}
    \centering
    \includegraphics[width=0.99\textwidth]{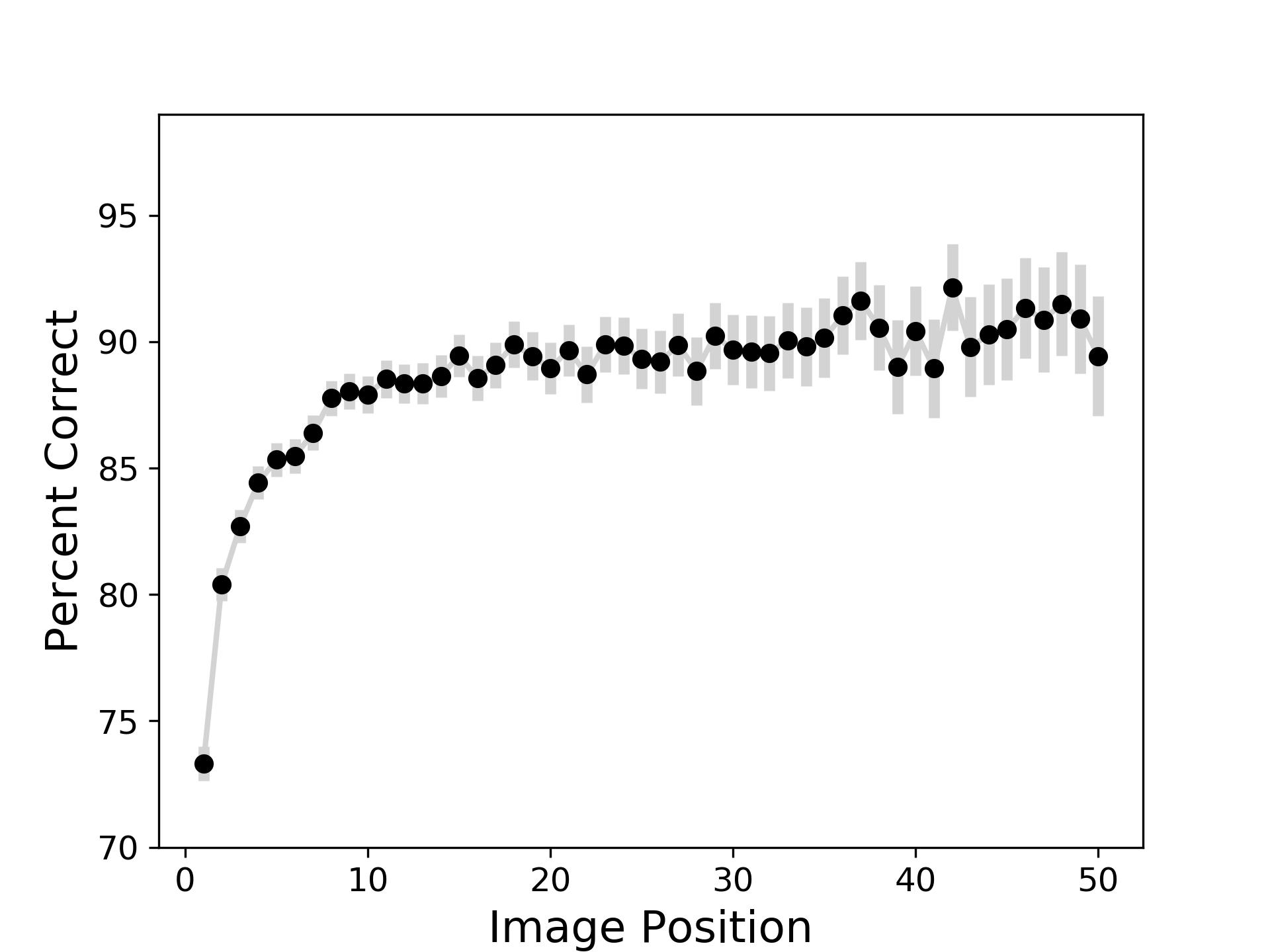}
    \caption{}
\end{subfigure}%
~
\begin{subfigure}[t]{0.49\textwidth}
    \centering
    \includegraphics[width=0.99\textwidth]{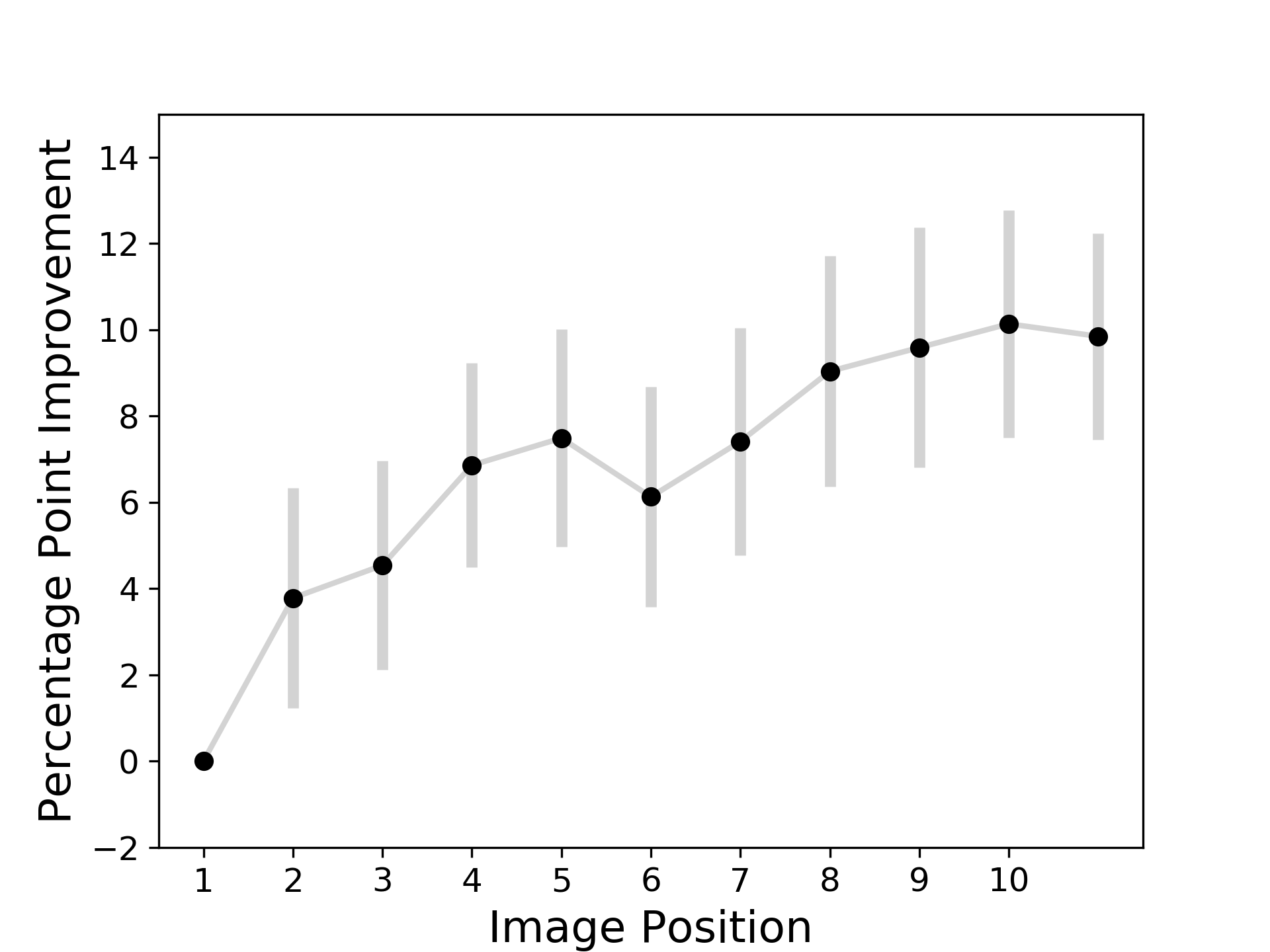}
    \caption{}
\end{subfigure}%
\caption{Participants' overall and marginal accuracy by image order with error bars showing a 95\% confidence interval for each image position -- (a) overall accuracy for all users with no fixed effects (b) marginal accuracy (relative to the first image position) for all users who saw at least 10 images controlling for user and image fixed effects and clustering errors at the image level. In (b), the 11th position includes all images positions beyond the 10th.}
\label{fig:regression_plot}%
\end{figure*}

Across all four robustness checks with and without fixed-effects, our models show a positive and statistically significant relationship between $T_{n}$ and $\hat{y}_{i,j}$. In the linear-log model, a one unit increase in $\log(T_{i_n})$ is associated with a 3 percentage point increase in $\hat{y}_{i,j}$. This effect is significant at the p<.01 level. In the model that estimates Equation \ref{equation:two}, we find a 1 percentage point average marginal treatment effect size of image position on $\hat{y}_{i,j}$. This effect is also significant at the p<.01 level. In other words, users improve their ability to guess by 1 percentage point for each of the first 10 guesses. Figure \ref{fig:regression_plot} shows these results graphically.

The statistically significant improvement in accurately identifying manipulations suggests that within the context of Deep Angel, exposure to media manipulation and feedback on what has been manipulated can successfully prepare individuals to detect faked media. When trained for 1 minute and 14 seconds on average, across ten images, participants improved their ability to detect manipulations by ten percentage points. As participants are exposed to image manipulations, they quickly learn to spot the vast majority of the manipulations. An examination of participants' learning rate across a variety of image and participant characteristics provides some insight into what drives learning.

\begin{figure*}[t!]
\centering

\includegraphics[width=0.16\linewidth]{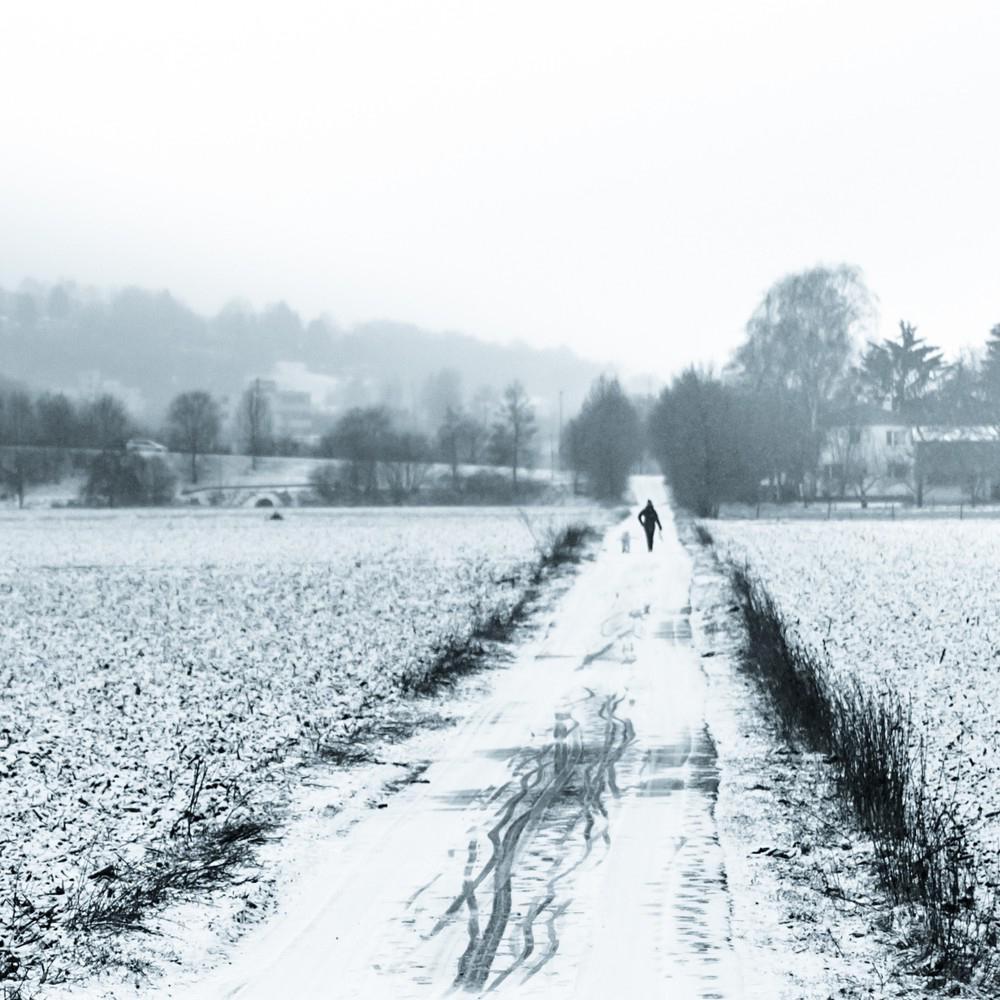}
\includegraphics[width=0.16\linewidth]{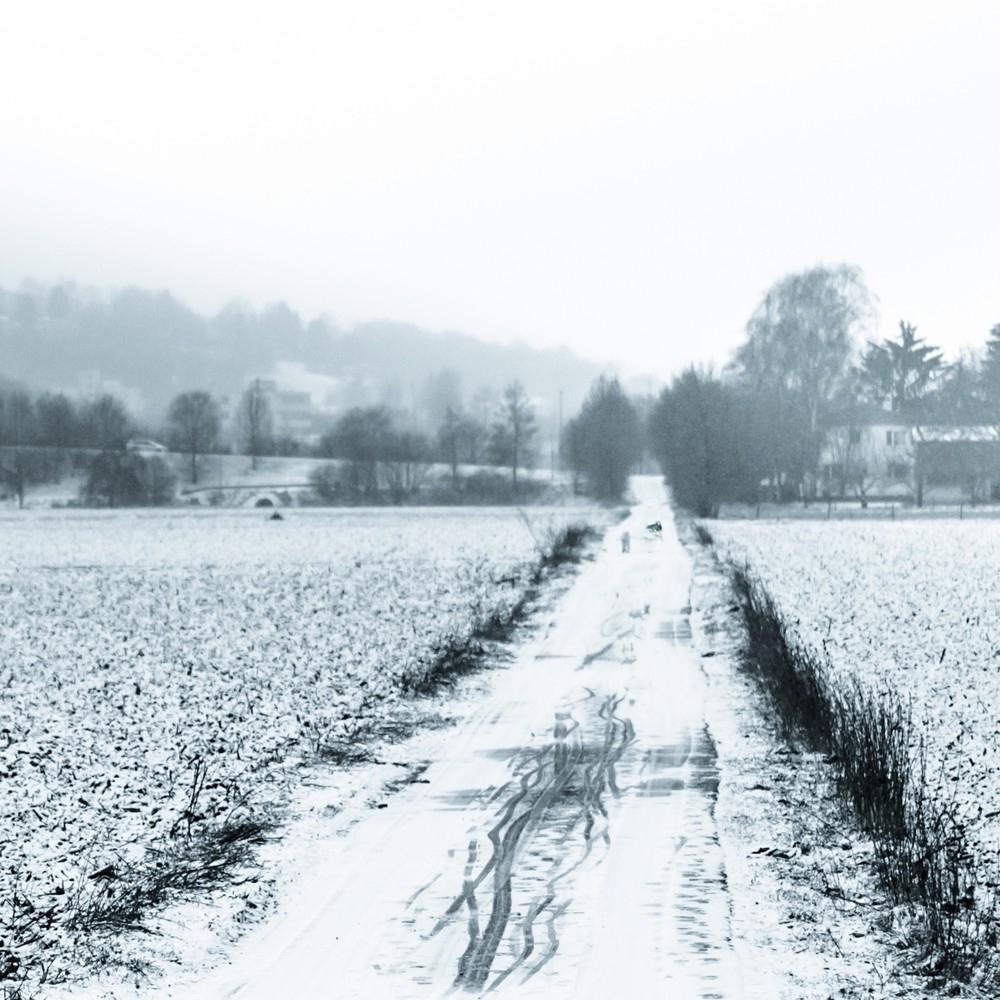}
\includegraphics[width=0.16\linewidth]{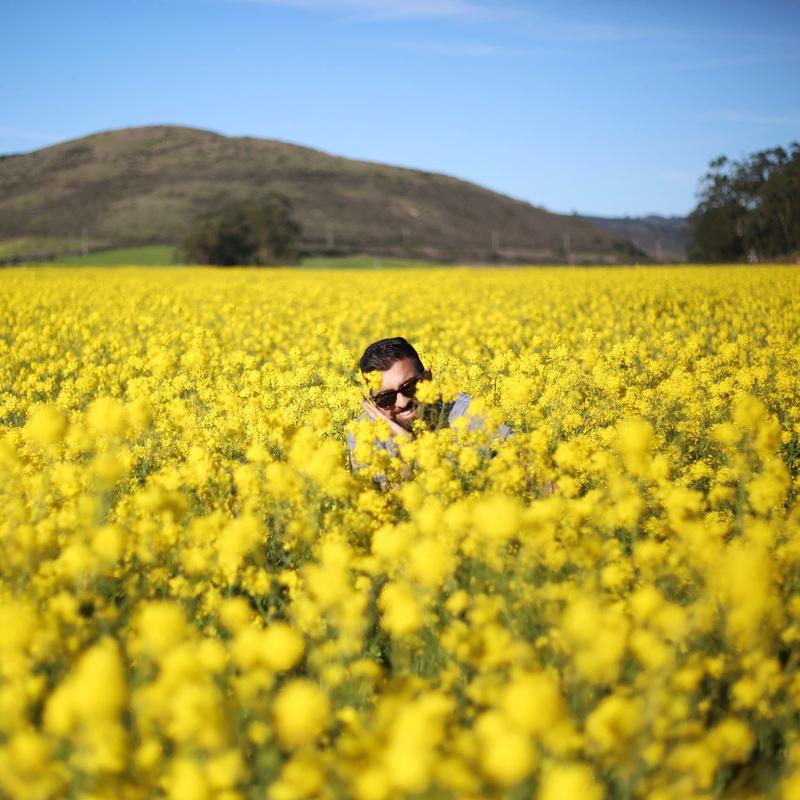}
\includegraphics[width=0.16\linewidth]{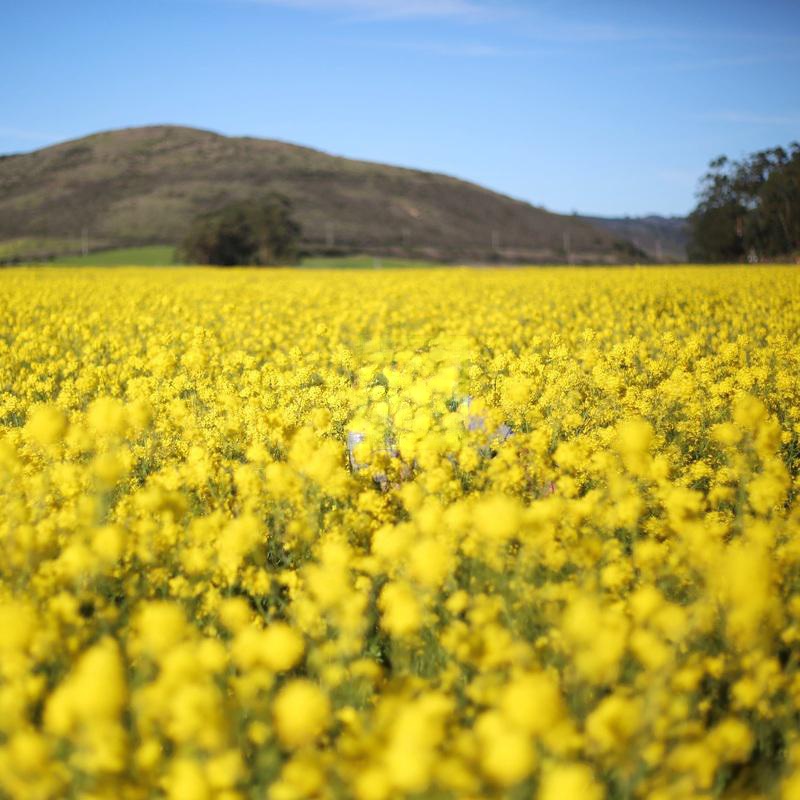}
\includegraphics[width=0.16\linewidth]{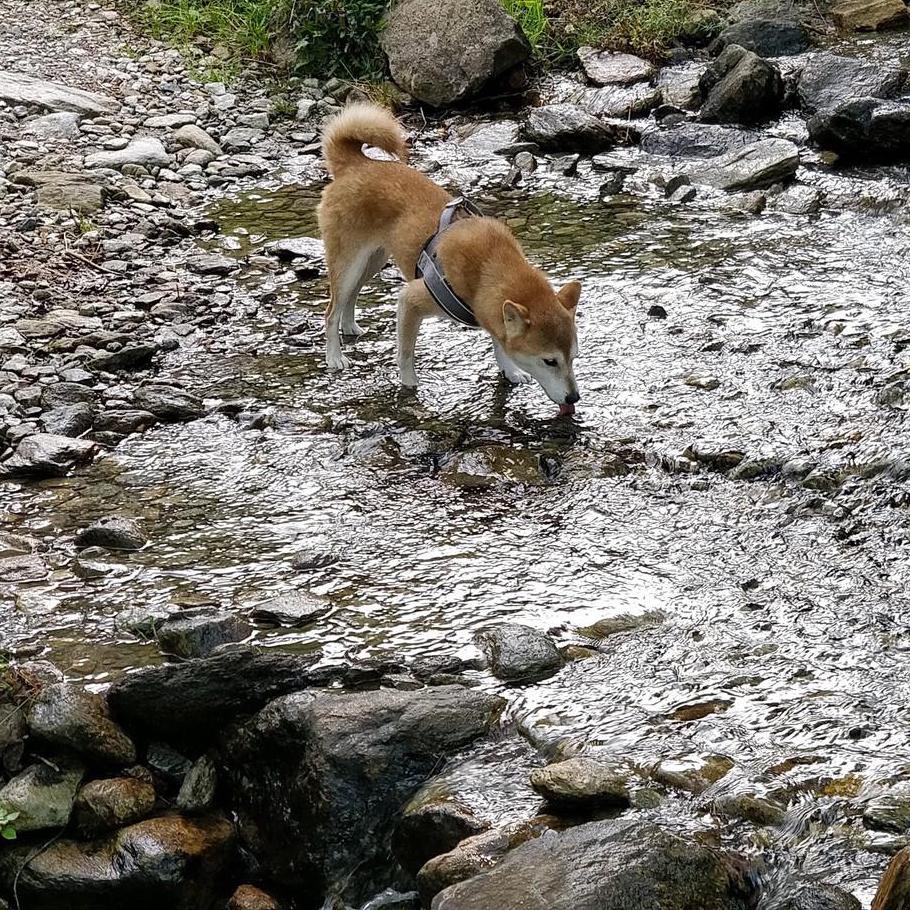}
\includegraphics[width=0.16\linewidth]{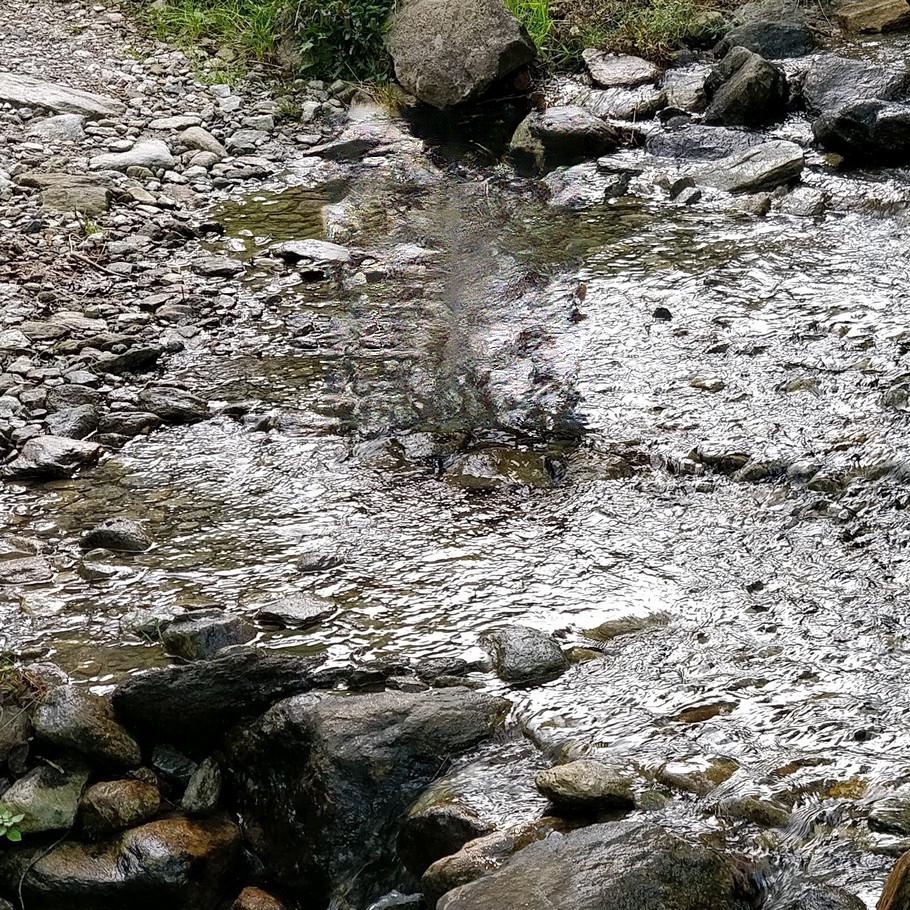}

\begin{subfigure}[t]{0.24\textwidth}
    \centering
    \includegraphics[width=0.99\textwidth]{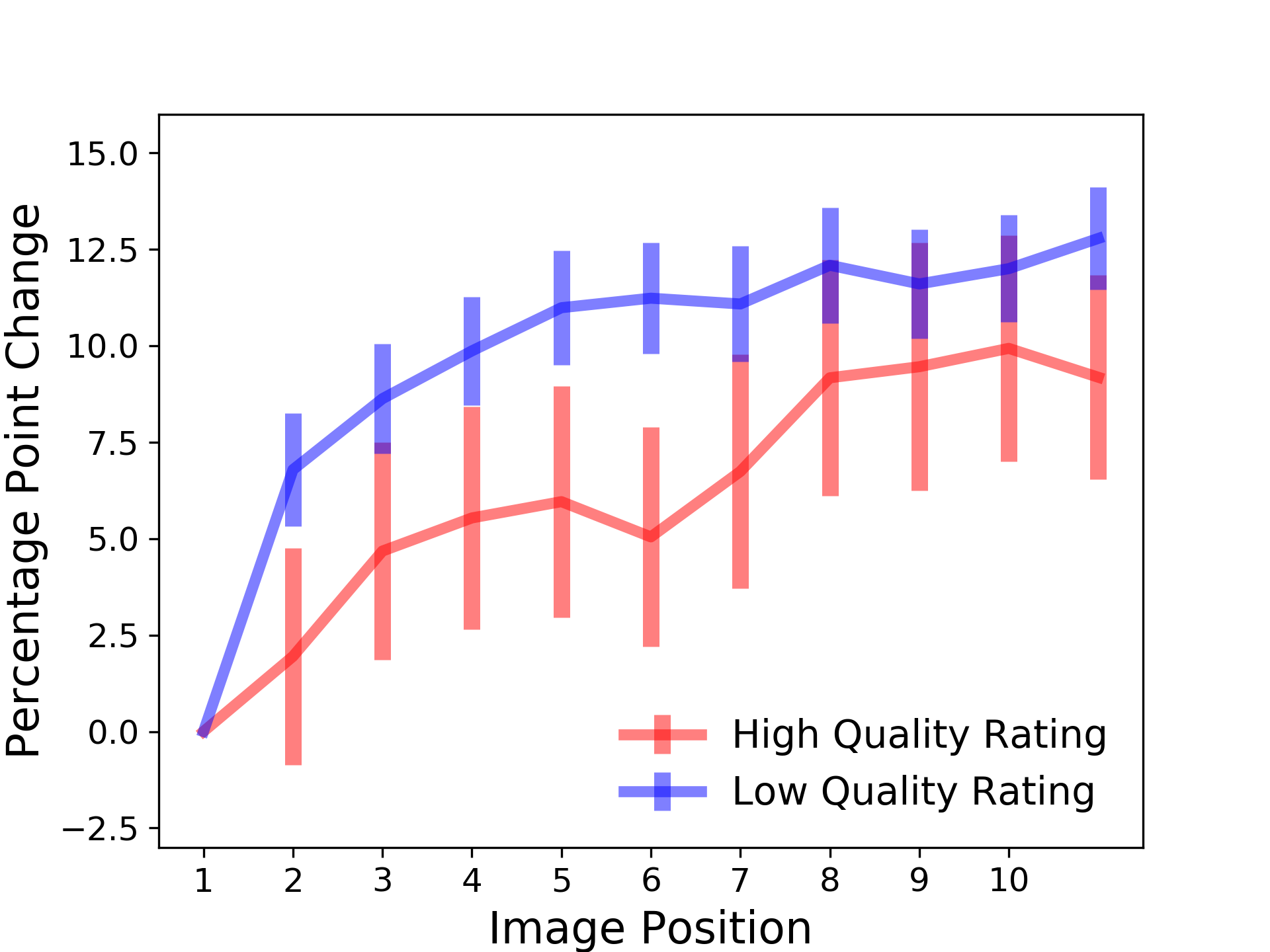}
    \caption{Subjective Quality}
\end{subfigure}%
~
\begin{subfigure}[t]{0.24\textwidth}
    \centering
    \includegraphics[width=0.99\textwidth]{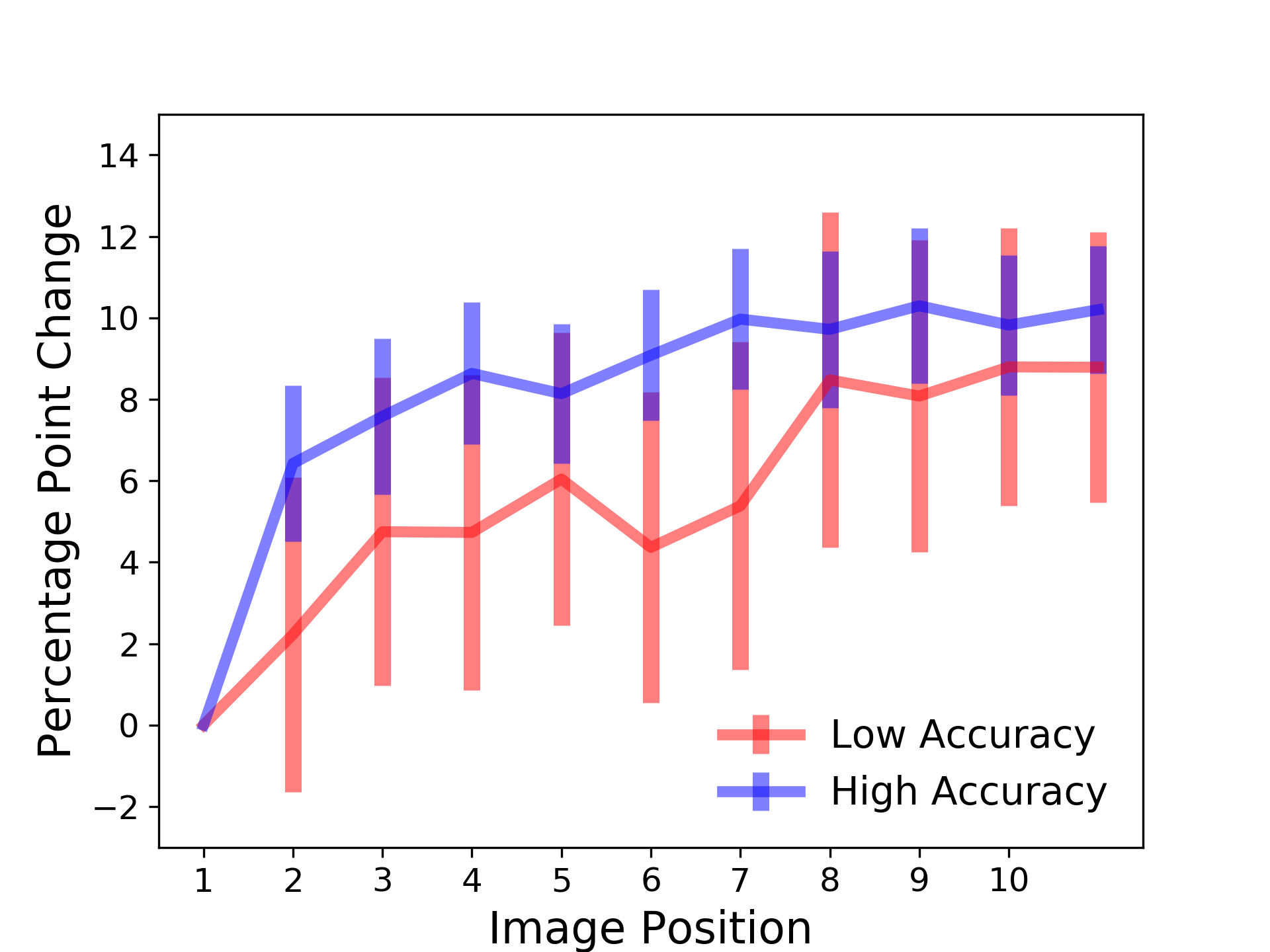}
    \caption{Image Accuracy}
\end{subfigure}%
~
\begin{subfigure}[t]{0.24\textwidth}
    \centering
    \includegraphics[width=0.99\textwidth]{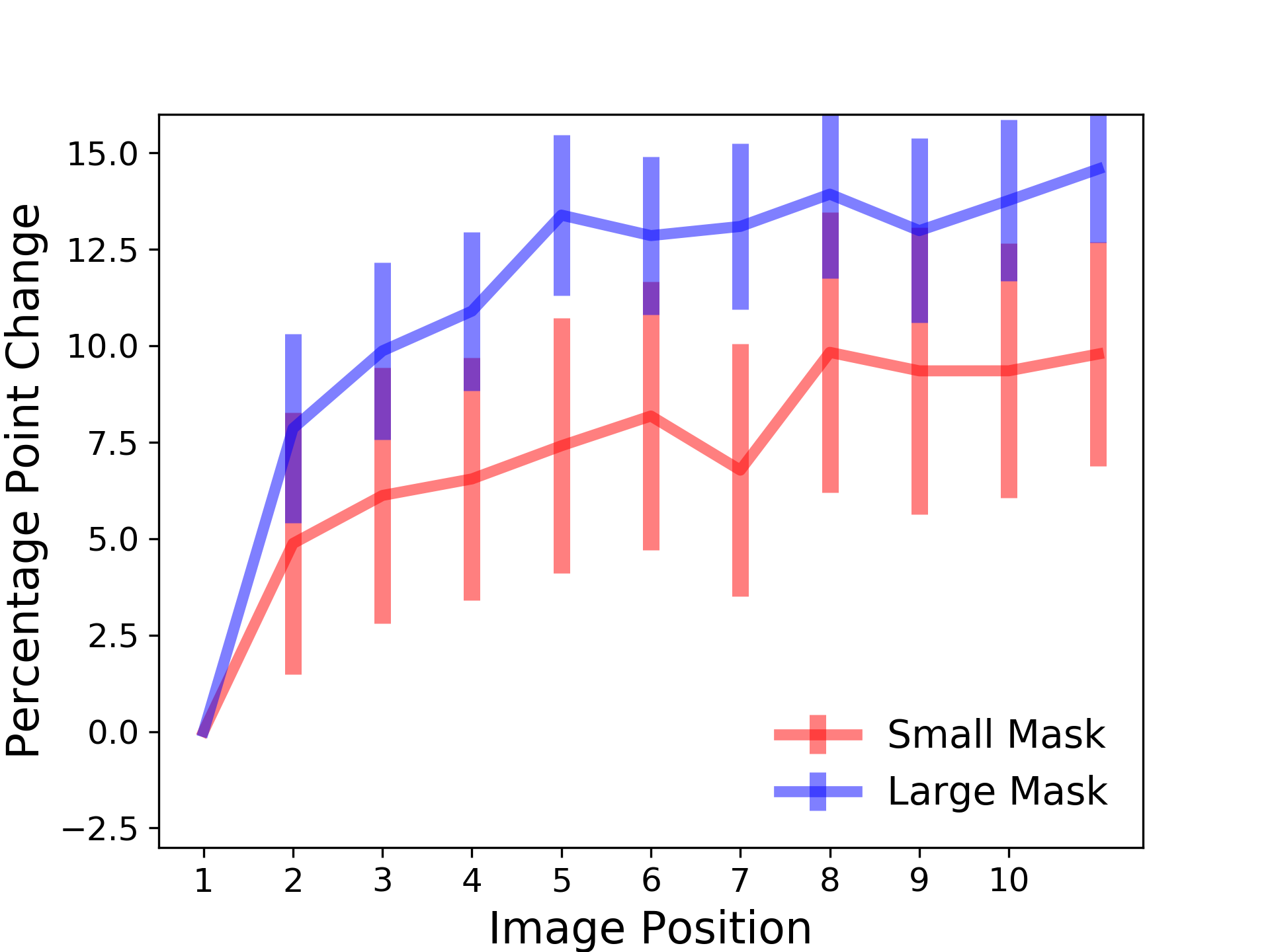}
    \caption{Mask Size}
\end{subfigure}%
~
\begin{subfigure}[t]{0.24\textwidth}
    \centering
    \includegraphics[width=0.99\textwidth]{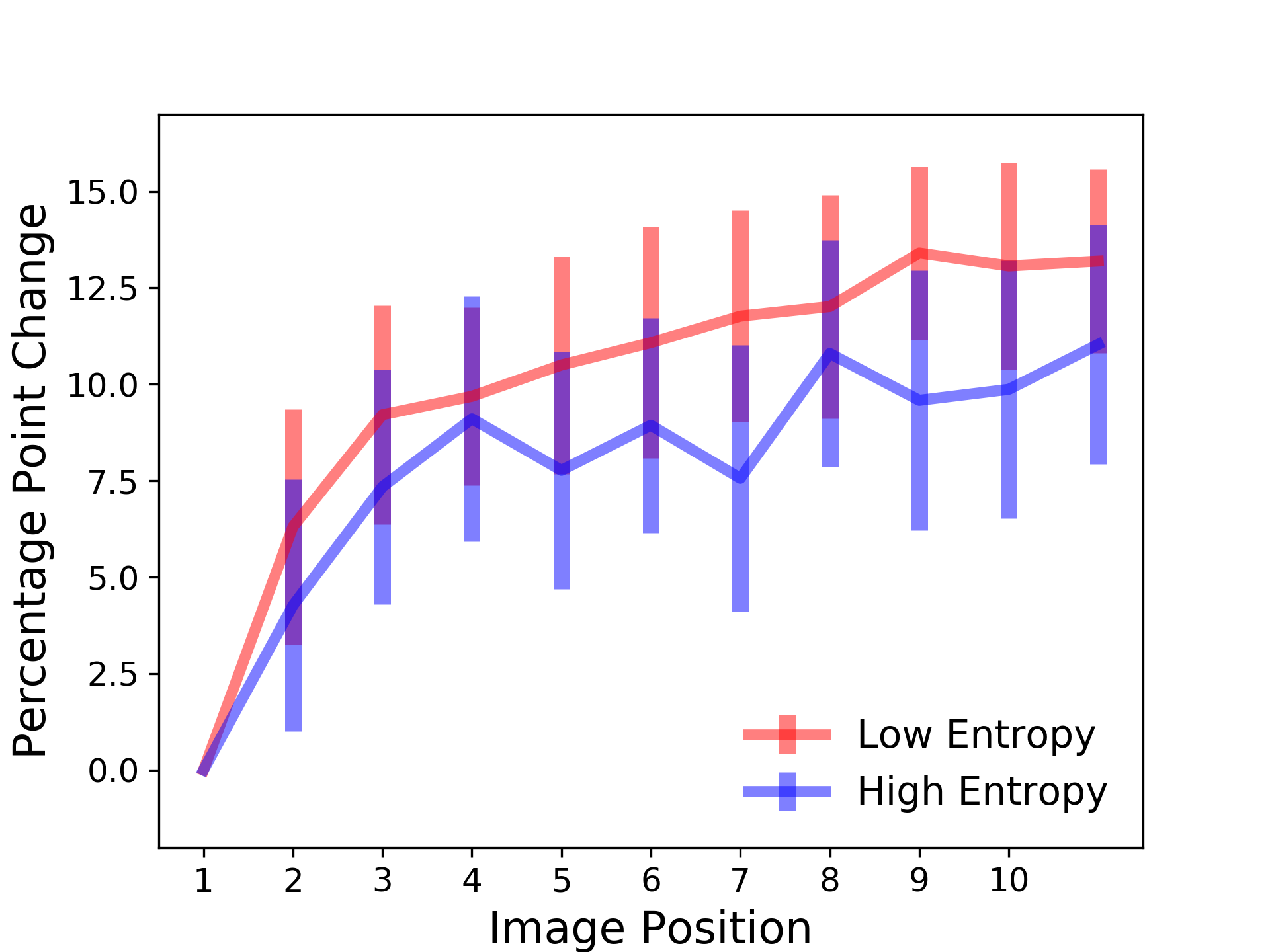}
    \caption{Entropy}
\end{subfigure}%

\begin{subfigure}[t]{0.24\textwidth}
    \centering
    \includegraphics[width=0.99\textwidth]{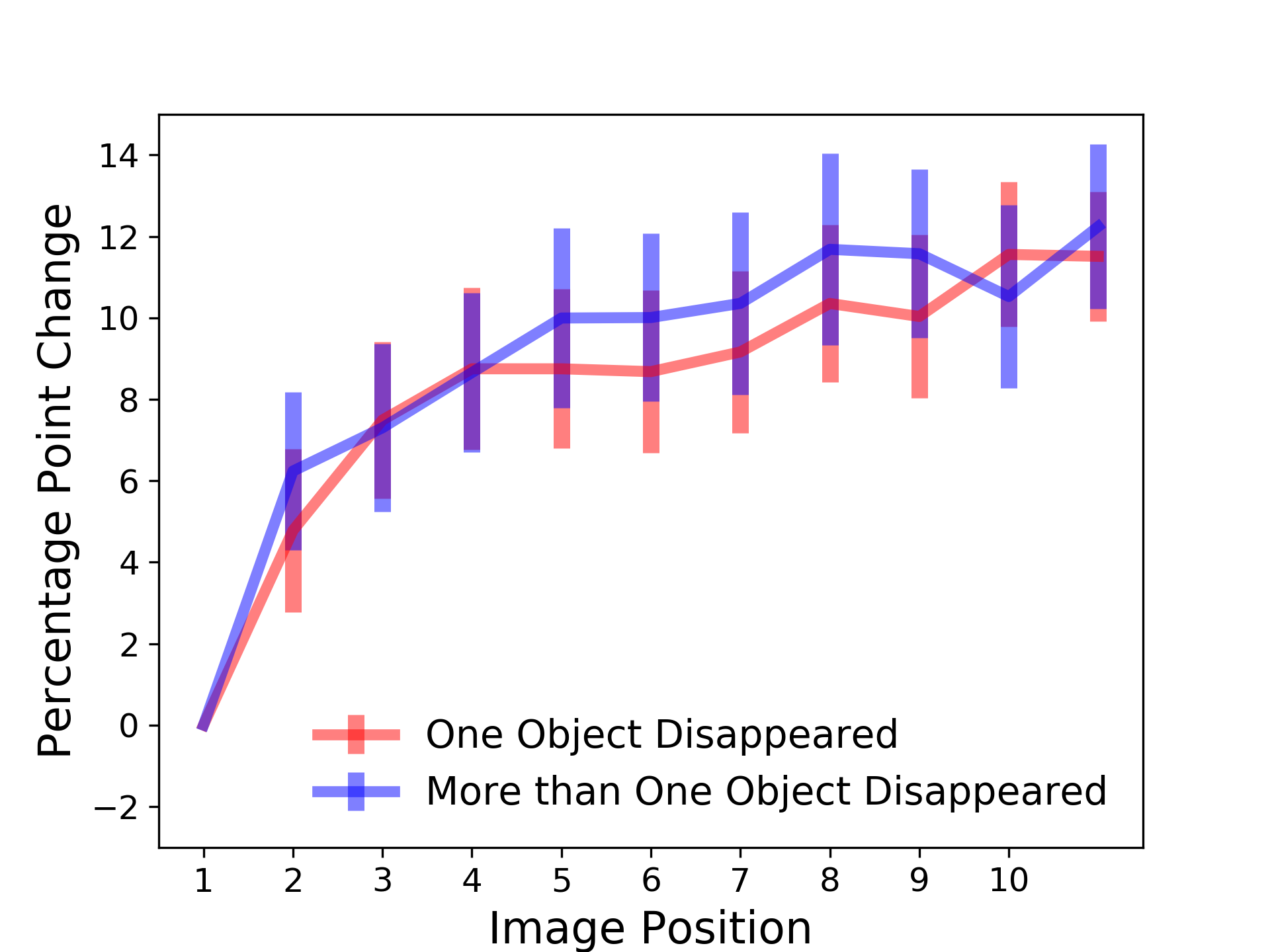}
    \caption{\# of Objects Disappeared}
\end{subfigure}%
~
\begin{subfigure}[t]{0.24\textwidth}
    \centering
    \includegraphics[width=0.99\textwidth]{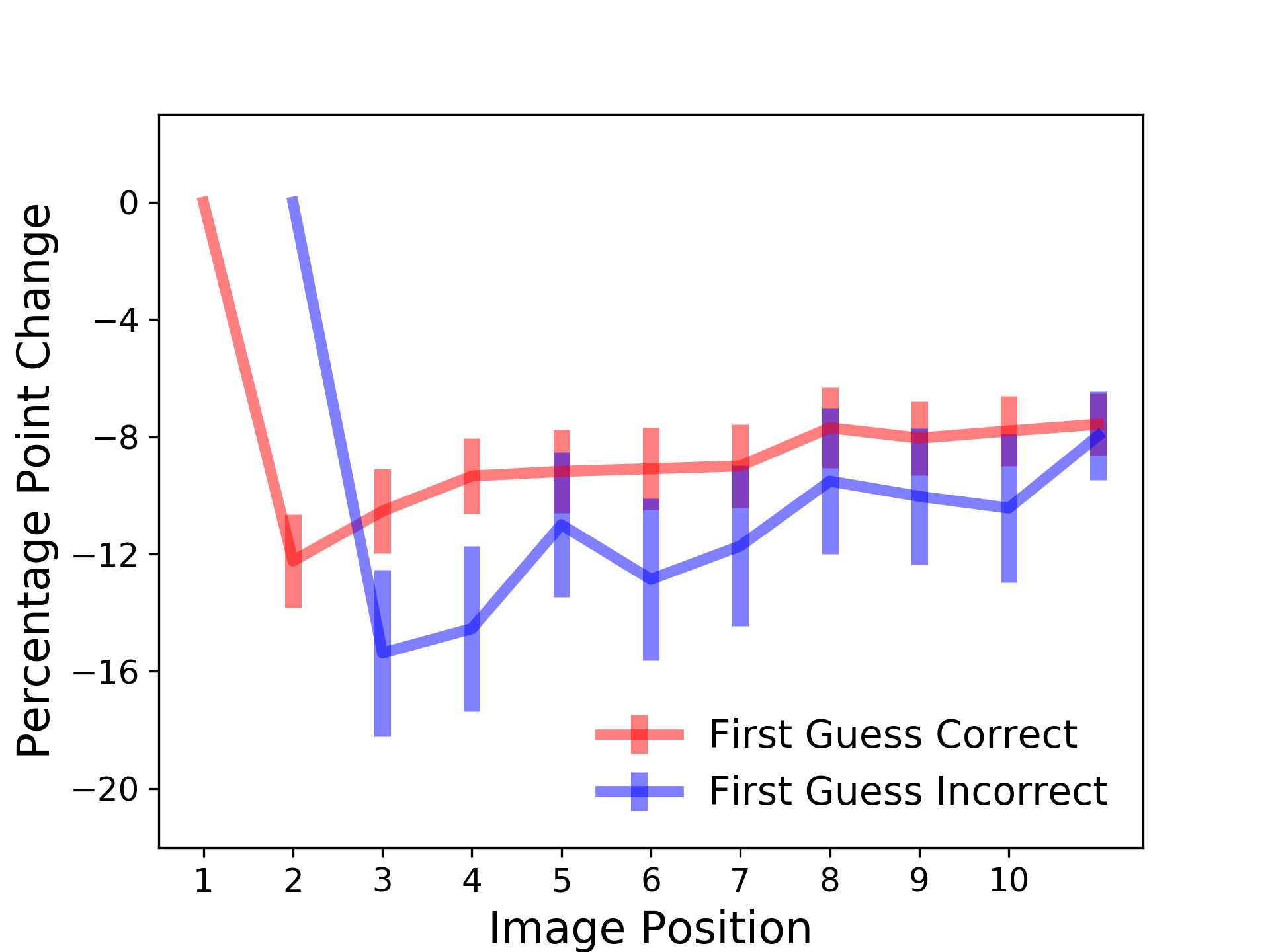}
    \caption{First Guess}
\end{subfigure}%
~
\begin{subfigure}[t]{0.24\textwidth}
    \centering
    \includegraphics[width=0.99\textwidth]{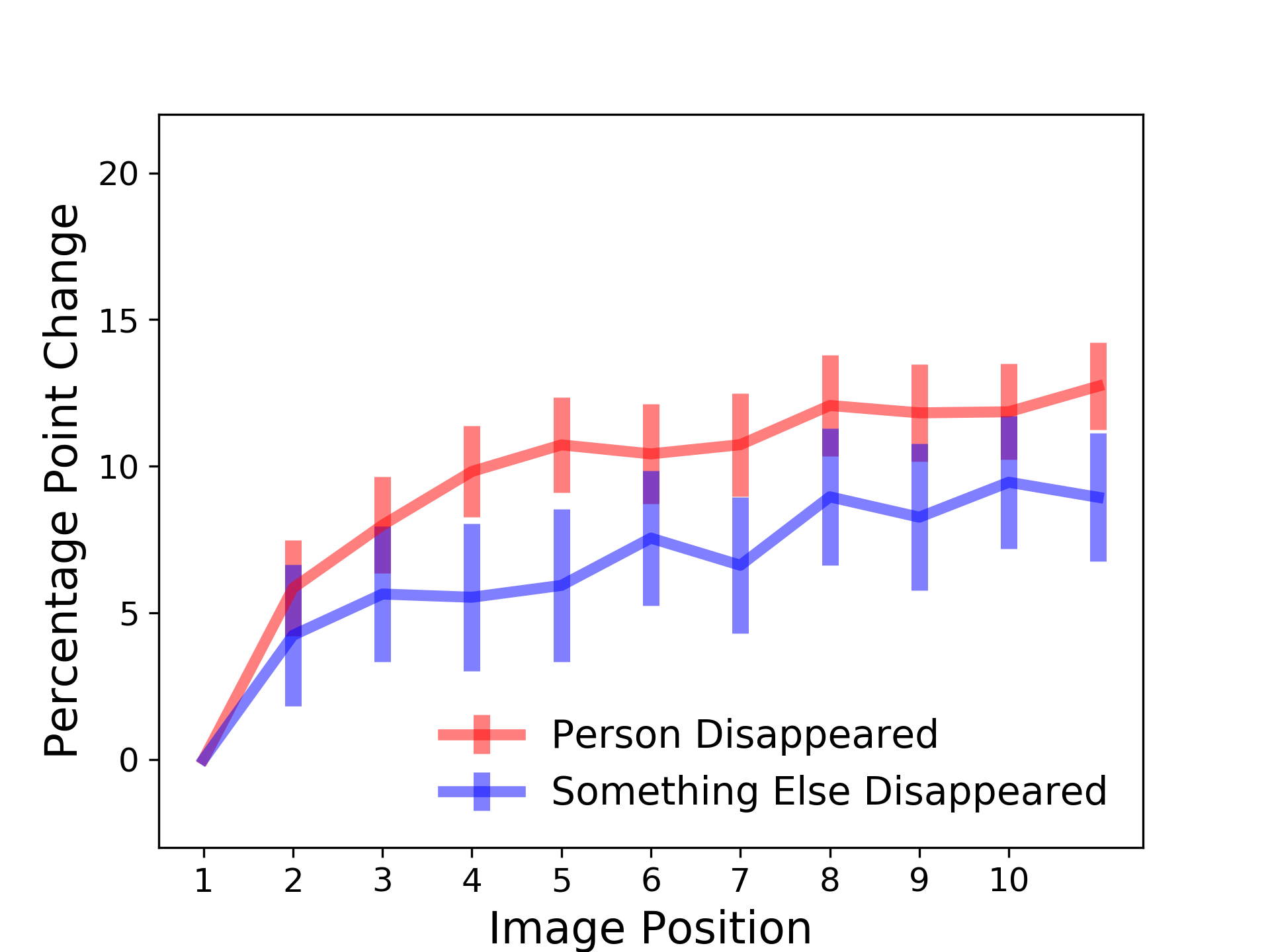}
    \caption{Has Person}
\end{subfigure}%
~
\begin{subfigure}[t]{0.24\textwidth}
    \centering
    \includegraphics[width=0.99\textwidth]{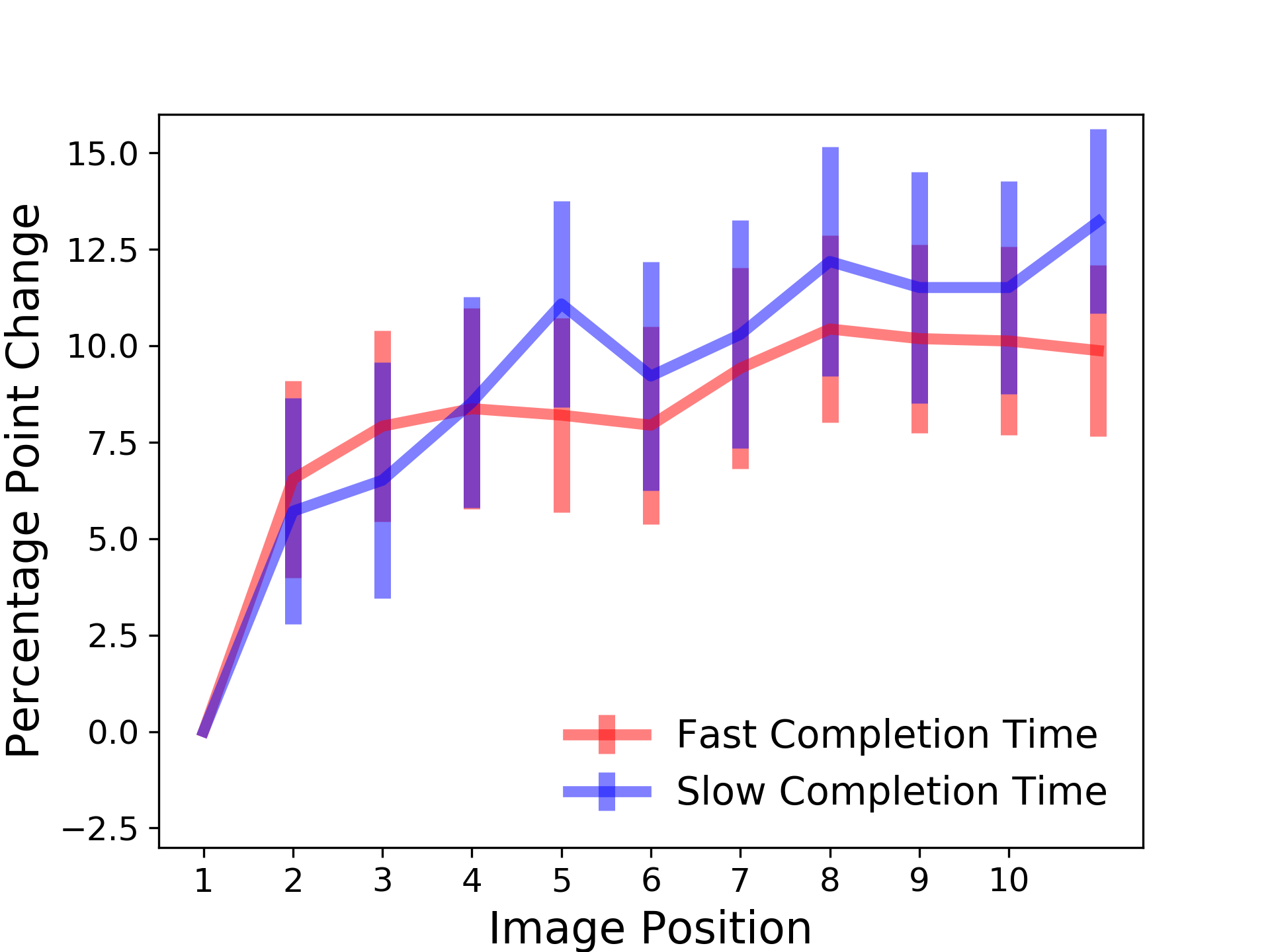}
    \caption{Completion Time}
\end{subfigure}%

\begin{subfigure}[t]{0.24\textwidth}
    \centering
    \includegraphics[width=0.99\textwidth]{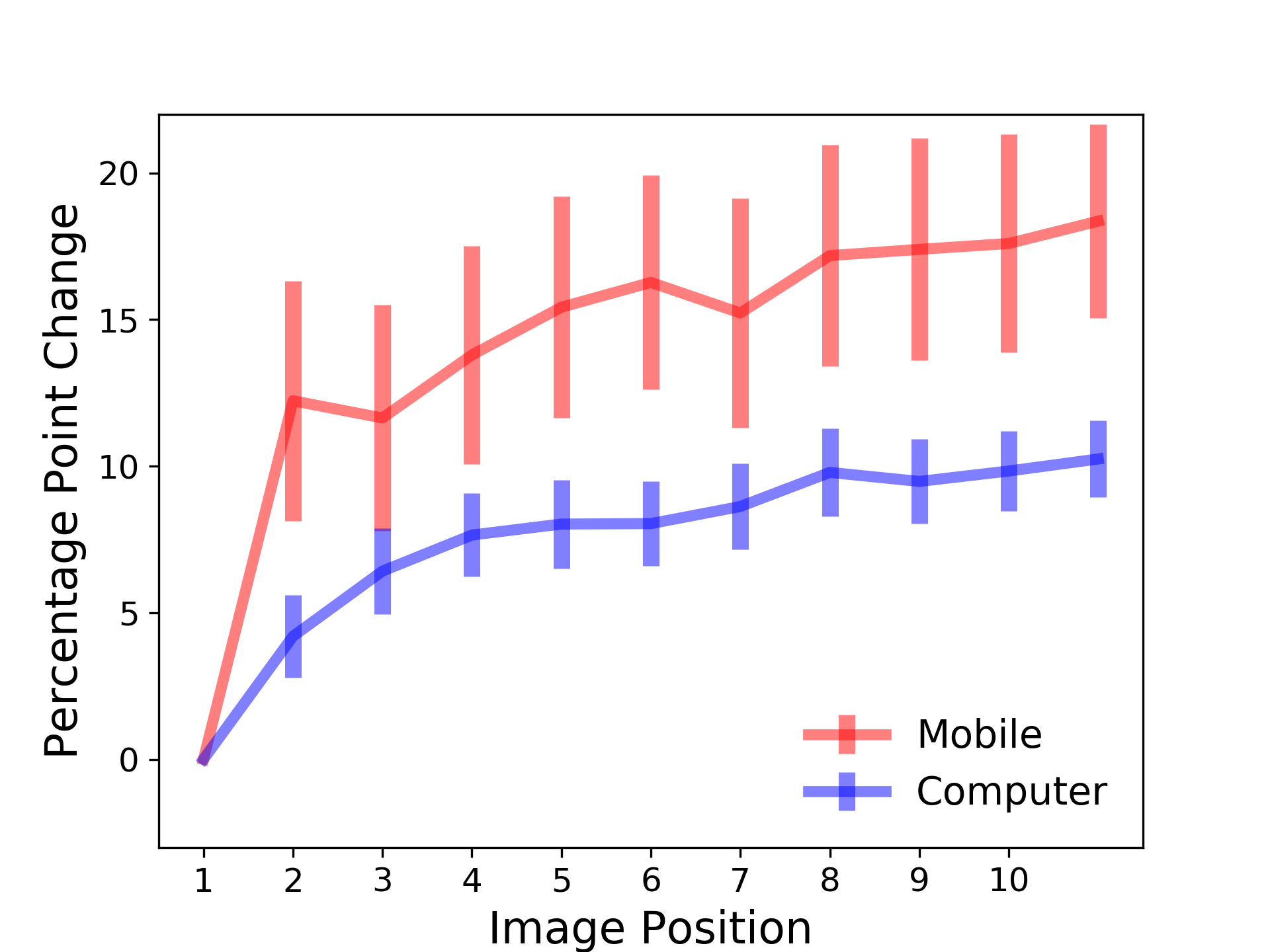}
    \caption{Mobile}
\end{subfigure}%
~
\begin{subfigure}[t]{0.24\textwidth}
    \centering
    \includegraphics[width=0.99\textwidth]{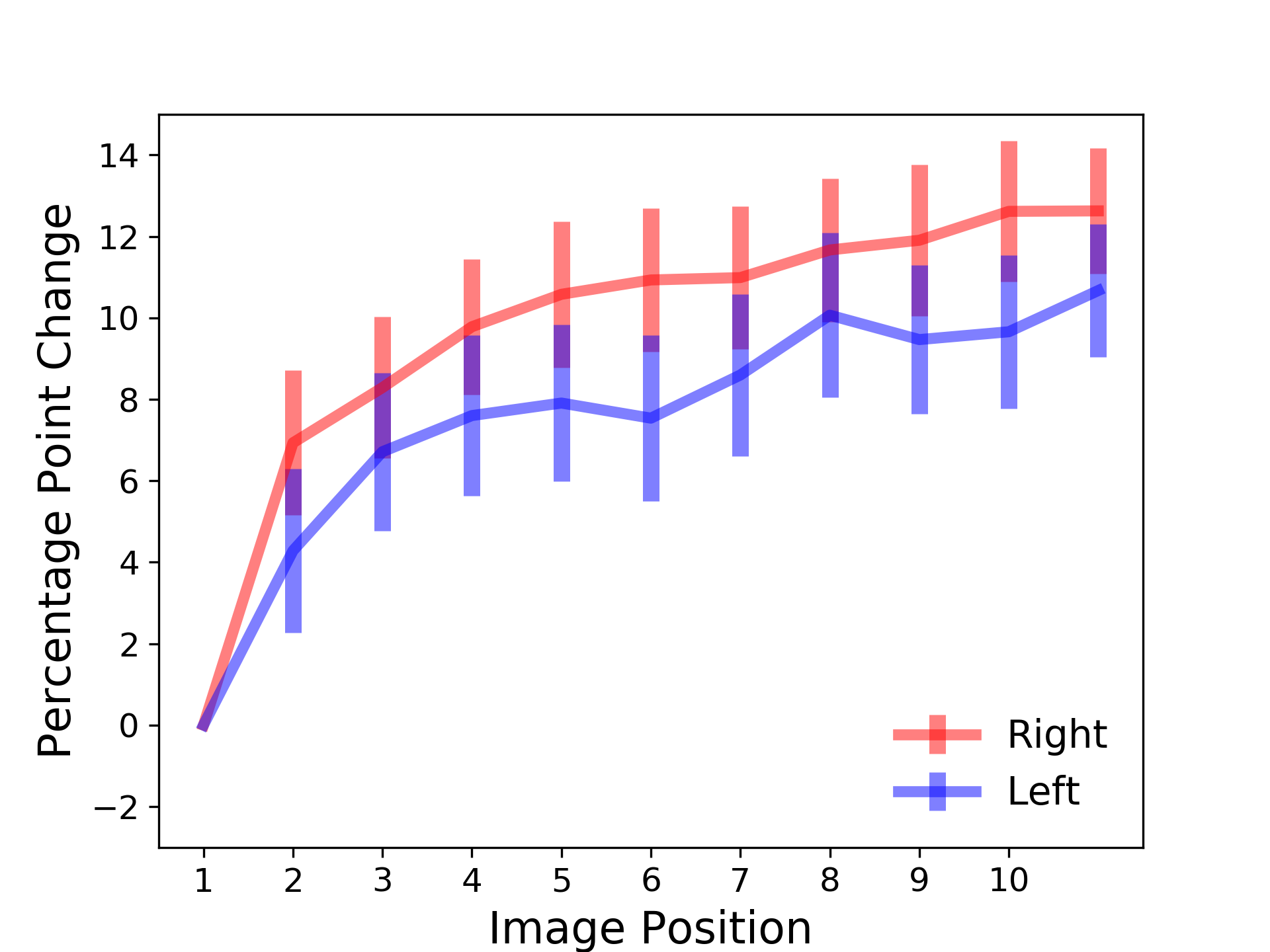}
    \caption{Image Location}
\end{subfigure}%

\caption{(i) From left to right, images (cropped into squares for display purposes) are increasing in entropy (3.5, 6.0, and 8.5), varying in percent of the image transformed (0.7\%, 2.4\%, and 1.9\%), similar in accuracy on the first guess (70\%, 71\%, 70\%), varying in accuracy beyond the first guess (88\%, 82\%, 92\%). (ii) 10 plots displaying heterogeneous effects of image and participant characteristics on learning while controlling for user and image fixed effects (a) whether the subjective image quality was judged as high by a third party,  (b) whether the original image was in the 1st to 25th percentile of accuracy or 75th to 99th, (c) whether the original image was in the 1st to 25th percentile of image mask proportion or 75th to 99th, (d) whether the original image was in the 1st to 25th percentile of entropy or 75th to 99th, (e) whether there were one or multiple objects disappeared (f) whether the participant's first answer was correct (the omitted position for each learning curve represents perfect accuracy), (g) whether the image contained a person, (h) whether the original image was in the 1st to 25th percentile of time to evaluate 10 images or 75th to 99th, (i) whether the participant viewed the images on a mobile device or computer (j) whether the image was placed on the left or right side of the screen. The error bars represent the 95\% confidence interval for each image position and errors are clustered at the image level.}
\label{fig:hetero}%
\end{figure*}

We evaluate manipulation quality across five measures: (a) a subjective quality rating, (b) 1st and 4th quartile image entropy, (c) 1st and 4th quartile proportion of  area of the image that was manipulated, (d) 1st and 4th quartile mean identification accuracy per image, and (e) number of objects disappeared. Retrospectively, we hired a third party to rate each image's manipulation as high or low quality based on whether large and noticeable artifacts were created by the image manipulation, which we use as a subjective quality rating. The image entropy is measured based on delentropy, an extension of Shannon entropy for images.\cite{larkin2016reflections} As an example, Figure~\ref{fig:hetero} presents three pairs of images subjectively rated as high quality with their corresponding entropy scores, proportion of the image transformed, mean accuracy of participants' first guess, and mean accuracy of participants guesses beyond the first to offer an example of what study participants learned.

The mean image accuracy of images subjectively rated as high quality is 75\% and 83\% for the 1st and 10th image, respectively, while the accuracy for the low quality images is 82\% and 94\%, respectively. While, Table~\ref{table:ols_3} shows that the difference in means across the subjective quality measure is statistically significant at the 99\% confidence level (p<.01), we do not find a statistically significant difference in learning rates. As shown in Figure~\ref{fig:hetero}a, there is a slight overlap in the confidence intervals of image positions 2 through 5, a statistically significant difference in image positions 6 and 7, and overlap in the confidence intervals for the rest of the image positions. There is some evidence that participants learn to identify low quality images faster than high quality images but the lack of statistically significant results on 8 of the 10 image positions and the lack of statistically significant results in the interaction between subjective quality and the logarithm of the image position in Table~\ref{table:ols_3} precludes us from rejecting the null hypothesis that the learning rate is identical across high and low quality images. These results indicate that the main effect is not simply driven by participants becoming proficient at guessing low-quality images in our data. 

In addition to the subjective quality rating, we evaluate the learning rate across four additional proxies for manipulation quality. Participants learn to identify low entropy images faster than high entropy images and images with large masked area faster than images with small masked area. Table~\ref{table:ols_3} shows that this difference in learning rates is statistically significant at the 95\% (p<.05) and 90\% (p<.10) levels, respectively. Smaller masked areas and lower entropy is associated with less stark and more subtle changes between an original and manipulated image. This relationship may indicate that participants learn more from subtle images than more obvious manipulations. Neither the split between the 1st and 4th quartile of mean accuracy per image nor the split between one object and many objects disappeared has a statistically significant effect on the learning rates.

We find heterogeneous effects on the learning rate based on participants' initial performance. In Figure~\ref{fig:hetero}, we compare subsequent learning rates of participants who correctly identified a manipulation on their first attempt to participants who failed on their first attempt and succeeded on their second. In this comparison, the omitted position for each learning curve represents perfect accuracy, which makes the marginal effects of subsequent image positions negative relative to these omitted image positions. On the first 3 of 4 image positions in this comparison, which correspond to the 3rd through 6th image positions, we find that initially successful participants learn faster than participants who were initially unsuccessful. This heterogeneous effect does not persist in the 7th position or beyond. Overall, this heterogeneous effect is statistically significant at the 99\% level (p<.01) and can be interpreted as initial performance is associated with a 2 percentage point reduction in the learning rate. This suggests that people who are more accurate on average also learn faster on average.

Participants learn to detect images with people disappeared faster than images with any other kind of thing disappeared. This difference is statistically significant at the 95\% confident interval (p<.05) in the log-linear regression as shown in Table~\ref{table:ols_3}. Figure~\ref{fig:hetero} also shows this difference is statistically significant in 2 of the 10 image positions. We conjecture that the photos where people have been disappeared induce a feeling that something is missing whereas when material objects like a soccer ball or cell phone are missing, the feeling of absence is not evoked. 

We do not find strong evidence that the speed with which a participant completed rating 11 images is related to the learning rate. We do not find evidence that the image placement is correlated with overall accuracy, but we do find statistically significant evidence at the 95\% confidence interval (p<.05) that placement of the image on the right is associated with a 1 percentage point increase in the learning rate, which may be related to how participants eyes scan the screen from left to right.

There is a clear difference in the learning rate of participants based on whether they participated with mobile phones or computers. Participants on mobile phones learn at a consistently faster rate than participants on computers, and this difference is statistically significant as shown in Table~\ref{table:ols_3} and displayed across 9 of 10 image positions in Figure~\ref{fig:hetero}. We conjecture that the ease of the zoom feature on mobile phones relative to computers enables mobile participants to inspect each image more closely.

\section*{Discussion}

While AI models can improve clinical diagnoses \cite{esteva2017dermatologist,poplin2018prediction, kooi2017large} and bring about autonomous driving \cite{chen2015deepdriving}, they also have the potential to scale censorship \cite{roberts2018censored}, amplify polarization \cite{Bakshy1130}, and spread both fake news \cite{Vosoughi1146} and manipulated media. We present results from a large scale randomized experiment that show the combination of exposure to manipulated media and feedback on what media has been manipulated improves individuals' ability to detect media manipulations. Direct interaction with cutting edge technologies for content creation might enable more discerning media consumption across society. In practice, the news media has exposed high-profile AI manipulated media including fake videos of the Speaker of the House of Representatives, Nancy Pelosi, and the CEO of Facebook, Mark Zuckerberg, which serves as feedback to everyone on what manipulations look like \cite{pelosi, zuck}. Our results build on recent research that suggests human intuition can be a reliable source of information about adversarial perturbations to images \cite{zhou2019humans} and recent research that provides evidence that familiarising people with how fake news is produced may confer cognitive immunity to people when they are later exposed to misinformation \cite{jon}. In addition, our results offer suggestive evidence for what drives learning to detect fake content. In this experiment, presenting participants with low entropy images with minor manipulations on mobile devices increased learning rates at statistically significant levels. When given feedback, participants appear to learn best from the most subtle manipulations.

The generalizability of our results is limited to the images produced by our AI model, and a promising avenue for future research could expand the domains and models studied. Likewise, future research could explore to what degree individuals' ability to adaptively detect manipulated media comes from learning-by-doing, direct feedback, and awareness that anything is manipulated at all.

Our results suggest a need to re-examine the precautionary principle that is commonly applied to content generation technologies. In 2018, Google published BigGAN, which can generate realistic appearing objects in images, but while they hosted the generator for anyone to explore, they explicitly withheld the discriminator for their model \cite{brock2018large}. Similarly, OpenAI restricted access to their GPT-2 model, which can generate plausible long-form stories given an initial text prompt, by only providing a pared down model of GPT-2 trained with fewer parameters \cite{gpt2}. If exposure to manipulated content can vaccinate people from future manipulations, then censoring dissemination of AI research on content generation may prove harmful to society by leaving it unprepared for a future of ubiquitous AI-mediated content. 

\section*{Methods}

We engineered a \textit{Target Object Removal} pipeline to remove objects in images and replace those objects with a plausible background. We combine a convolutional neural network (CNN) trained to detect objects with a generative adversarial network (GAN) trained to inpaint missing pixels in an image \cite{goodfellow2014generative, karras2017progressive, DBLP:journals/corr/HeGDG17, DBLP:journals/nature/LeCunBH15}. Specifically, we generate object masks with a CNN based on a RoIAlign bilinear interpolation on nearby points in the feature map \cite{DBLP:journals/corr/HeGDG17}. We crop the object masks from the image and apply a generative inpainting architecture to fill in the object masks \cite{IizukaSIGGRAPH2017,yu2018generative}. The generative inpainting architecture is based on dilated CNNs with an adversarial loss function which allows the generative inpainting architecture to learn semantic information from large scale datasets and generate missing content that makes contextual sense in the masked portion of the image \cite{yu2018generative}.

\subsection*{Target Object Removal Pipeline}

\begin{figure*}[h]
\centering
\includegraphics[width=0.99\textwidth]{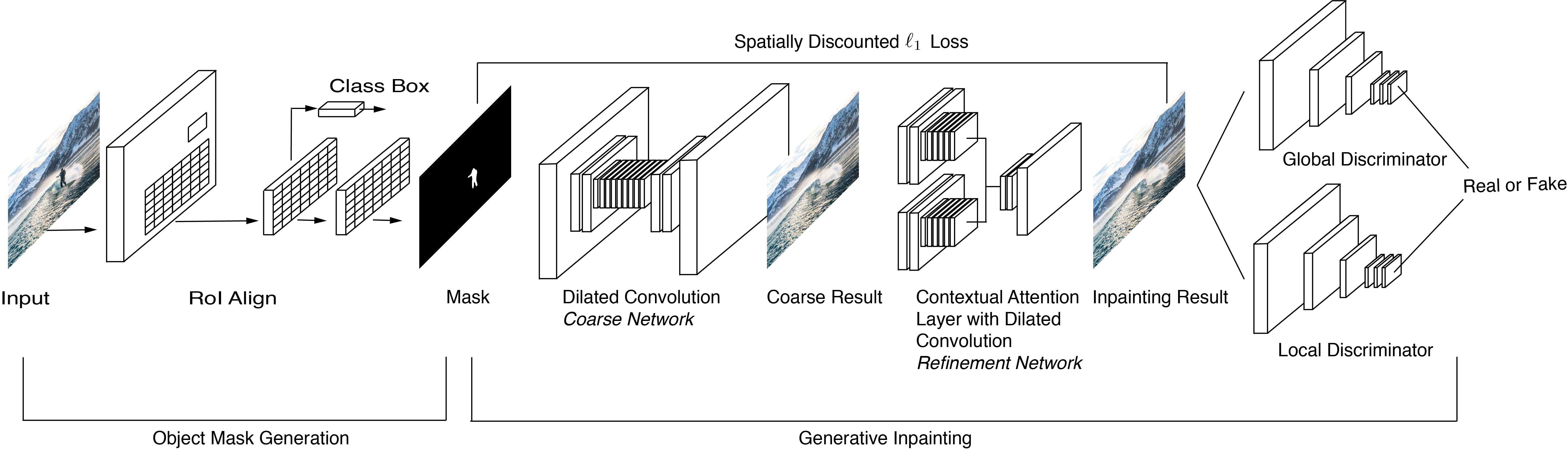}
\caption{End-to-end pipeline for targeted object removal following \cite{DBLP:journals/corr/HeGDG17,yu2018generative}}
\label{fig:neural_architecture}%
\end{figure*}

Our end-to-end \textit{targeted object removal} pipeline consists of three interfacing neural networks:
\begin{itemize}
\item \textbf{Object Mask Generator (G):} This network creates a segmentation mask $\hat{X} = G(X,y)$ given an input image $X$ and a target class $y$. In our experiments, we initialize \textbf{G} from a semantic segmentation network trained on the 2014 MS-COCO dataset following the Mask-RCNN algorithm \cite{DBLP:journals/corr/HeGDG17}. The network generates masks for all object classes present in an image, and we select only the correct masks based on input $y$. This network was trained on 60 object classes.
\item \textbf{Generative Inpainter (I):} This network creates an inpainted version $Z = I(\hat{X}, X)$ of the input image $X$ and the object mask $\hat{X}$. \textbf{I} is initialized following the DeepFill algorithm trained on the MIT Places 2 dataset~\cite{yu2018generative, zhou2017places}.
\item \textbf{Local Discriminator (D):} The final discriminator network takes in the inpainted image and determines the validity of the image. Following the training of a GAN discriminator, \textbf{D} is trained simultaneously on \textbf{I} where $X$ are images from the MIT Places 2 dataset and $\hat{X}$ are the same images with randomly assigned holes following~\cite{zhou2017places, yu2018generative}.
\end{itemize}
For every input image and class label pair, we first generate an object mask using \textbf{G}, which is paired with the image and inputted to the inpainting network \textbf{I} that produces the generated image. The inpainter is trained from the loss of the discriminator \textbf{D}, following the typical GAN pipeline. An illustration of our neural network architecture is provided in Figure \ref{fig:neural_architecture}.

\subsection*{Live Deployment}

\begin{figure}[ht]
    \centering
    \includegraphics[width=0.75\textwidth]{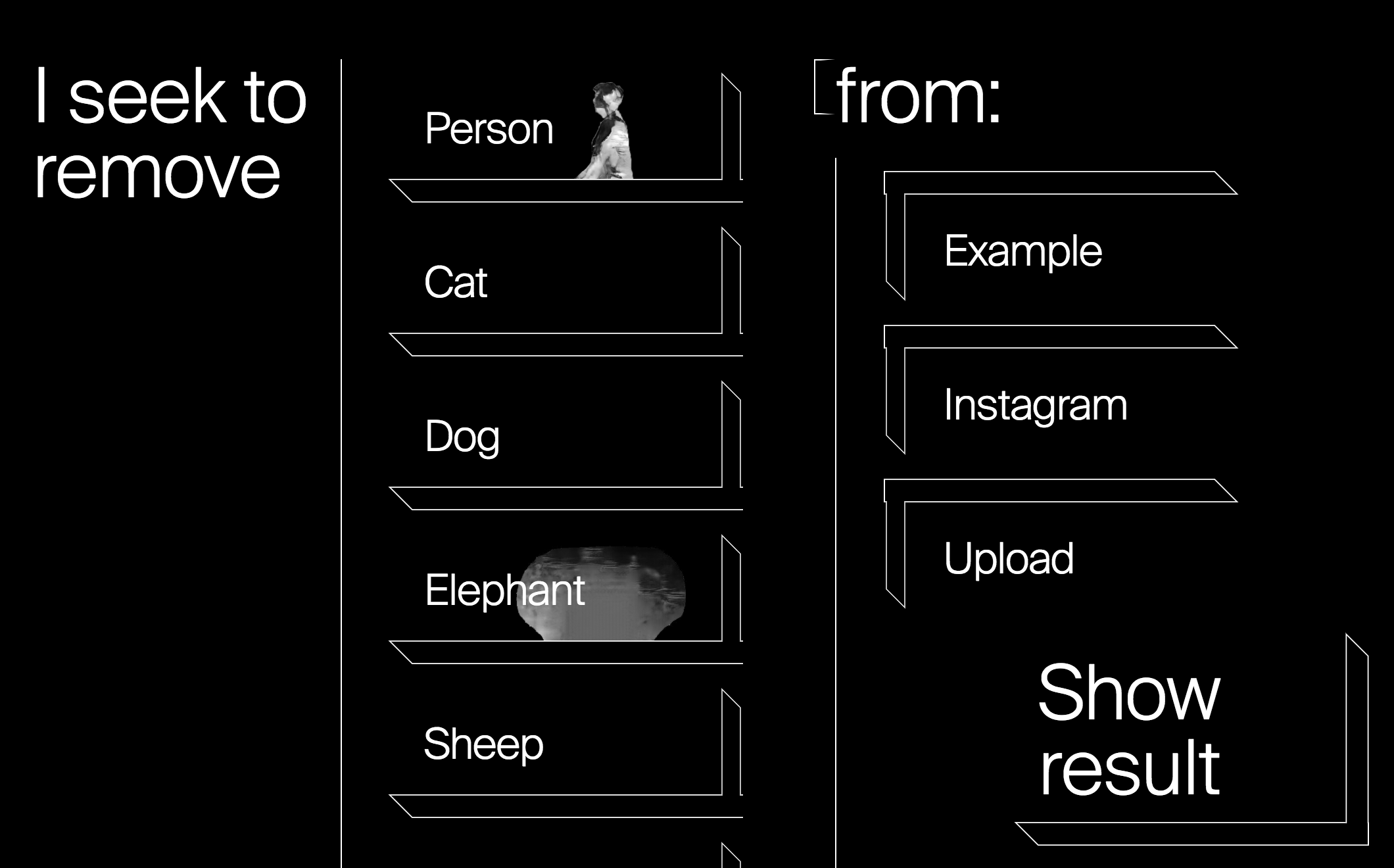}
    \caption{``Erase with AI'' User Interfaces}
    \label{fig:uierase}
\end{figure}

We designed an interactive website called Deep Angel to make the \textit{Target Object Removal} pipeline publicly available.\footnote{We retained the Cyberlaw Clinic from the Harvard Law School and Berkman Klein Center for Internet \& Society to advise and support us throughout the Deep Angel experiment.} The API for the \textit{Target Object Removal} pipeline is served by a single Nvidia Geforce GTX Titan X. In addition to the ``Detect Fakes'' user interaction, Deep Angel has a user interaction ``Erase with AI,'' where people can apply the \textit{Target Object Removal} pipeline on their own images. See Figure \ref{fig:uierase} for a screen shot of this user interface.

In ``Erase with AI,'' people first select a category of object that they seek to remove and then they either upload an image or select an Instagram account from which to upload the three most recent images. After the user submits his or her selections, Deep Angel returns both the original image and a transformation of the original image with the selected objects removed. 

Users uploaded 18,152 unique images from mobile phones and computers. In addition, user directed the crawling of 12,580 unique images from Instagram.  The most frequently selected objects for removal are displayed in Table SI\ref{table:selections}. 
The overwhelming majority of images uploaded and Instagram accounts selected were unique. 88\% of the usernames entered for targeted Instagram crawls were unique.

We can surface the most plausible object removal manipulations by examining the images with the lowest guessing accuracy.  Ultimately, plausible manipulations are relatively rare and image dependent. 

The \textit{Target Object Removal} model can produce plausible content but it is not perfect. For the \textit{Target Object Removal} model, plausible manipulations are confined to specific domains. Objects are only plausibly removed when they are a small portion of the image and the background is natural and uncluttered by other objects. Likewise, the model often generates model-specific artifacts that humans can learn to detect.

\noindent\textbf{Data Availability}: Upon publication, the data and replication code will be made available in a public Github repository.

\noindent\textbf{Acknowledgments}:  We thank Abhimanyu Dubey, Mohit Tiwari, and David McKenzie for their helpful comments
and feedback. 

\noindent\textbf{Author contributions}: M.G. implemented the methods, M.G., Z.E., N.O. analyzed data and wrote the paper. All authors conceived the original idea, designed the research, and provided critical feedback on the analysis and manuscript.

\bibliography{main}
\bibliographystyle{ieeetr}
\section*{Appendix I: Regression Tables}

\begin{table}[ht]
\begin{center}
\begin{tabular}{lcccc}
\hline
              &    (1)     &    (2)     &    (3)     &    (4)      \\
\midrule
\midrule
Log(Image Position)     & 0.0261***  & 0.0259***  & 0.0259***  & 0.0255***   \\
             & (0.0012)   & (0.0012)   & (0.0013)   & (0.0029)    \\
\hline
\\
\hline

$N$           & 242216     & 192665     & 172434     & 55692       \\
Mean Accuracy on $1^{st}$ Image & 0.73 & 0.78 & 0.78 & 0.74 \\
Mean Accuracy on $10^{th}$ Image & 0.88 & 0.88 & 0.88 & 0.83 \\
$R^2$       & 0.29       & 0.19       & 0.20       & 0.26        \\
\hline
\end{tabular}
\end{center}
\caption{Ordinary least squares regression with participant and image fixed effects evaluating image position on users' accuracy in identifying manipulated images. Robust standard errors clustered at the image level in parentheses. *, **, and *** indicates statistical significance at the 90, 95, and 99 percent confidence intervals, respectively. All columns include participant and image fixed effects. Column (1) includes all images (2) drops all users who submitted fewer than 10 guesses and removes all control images where nothing was removed (3) drops all observations where a user has already seen a particular image (4) keeps only the images qualitatively judged as very high quality.}
\label{table:ols_1}
\end{table}

\begin{table}[ht]
\begin{center}
\begin{tabular}{lcccc}
\hline
              &    (1)     &    (2)     &    (3)     &    (4)      \\
\midrule
\midrule
2nd          & 0.0507***  & 0.0569***  & 0.0571***  & 0.0378***   \\
             & (0.0042)   & (0.0059)   & (0.0060)   & (0.0131)    \\
3rd          & 0.0672***  & 0.0744***  & 0.0746***  & 0.0454***   \\
             & (0.0048)   & (0.0060)   & (0.0059)   & (0.0123)    \\
4th          & 0.0775***  & 0.0888***  & 0.0885***  & 0.0686***   \\
             & (0.0050)   & (0.0058)   & (0.0058)   & (0.0121)    \\
5th          & 0.0859***  & 0.0978***  & 0.0967***  & 0.0749***   \\
             & (0.0052)   & (0.0062)   & (0.0064)   & (0.0129)    \\
6th          & 0.0817***  & 0.0962***  & 0.0963***  & 0.0613***   \\
             & (0.0057)   & (0.0064)   & (0.0064)   & (0.0130)    \\
7th          & 0.0900***  & 0.1032***  & 0.1039***  & 0.0741***   \\
             & (0.0056)   & (0.0064)   & (0.0065)   & (0.0134)    \\
8th          & 0.1019***  & 0.1120***  & 0.1106***  & 0.0904***   \\
             & (0.0055)   & (0.0065)   & (0.0065)   & (0.0137)    \\
9th          & 0.1028***  & 0.1136***  & 0.1134***  & 0.0959***   \\
             & (0.0055)   & (0.0063)   & (0.0063)   & (0.0142)    \\
10th         & 0.1030***  & 0.1135***  & 0.1123***  & 0.1014***   \\
             & (0.0056)   & (0.0062)   & (0.0064)   & (0.0135)    \\
More than 10 & 0.1106***  & 0.1215***  & 0.1197***  & 0.0985***   \\
             & (0.0051)   & (0.0059)   & (0.0059)   & (0.0122)    \\
\hline
\\
\hline
$N$            & 242216     & 192665     & 172434     & 55692       \\
Mean Accuracy on $1^{st}$ Image & 0.73 & 0.78 & 0.78 & 0.74 \\
Mean Accuracy on $10^{th}$ Image & 0.88 & 0.88 & 0.88 & 0.83 \\
$R^2$           & 0.29       & 0.20       & 0.20       & 0.26        \\
\hline
\end{tabular}
\end{center}
\caption{Ordinary least squares regression with participant and image fixed effects evaluating image position on users' accuracy in identifying manipulated images. Robust standard errors clustered at the image level in parentheses. *, **, and *** indicates statistical significance at the 90, 95, and 99 percent confidence intervals, respectively. All columns include participant and image fixed effects. Column (1) includes all images (2) drops all users who submitted fewer than 10 guesses and removes all control images where nothing was removed (3) drops all observations where a user has already seen a particular image (4) keeps only the images qualitatively judged as very high quality.}
\label{table:ols_2}
\end{table}

\begin{table}
\begin{center}
\scalebox{0.6}{
\begin{tabular}{lcccccccccc}
\hline
                                       &    (1)     &    (2)     &    (3)     &    (4)     &    (5)     &    (6)     &    (7)     &    (8)     &    (9)     &    (10)     \\
\midrule
\midrule
Log(Image Position)                              & 0.0477***  & 0.0386***  & 0.0565***  & 0.0401***  & 0.0460***  & 0.0505***  & 0.0367***  & 0.0492***  & 0.0410***  & 0.0406***   \\
                                       & (0.0029)   & (0.0034)   & (0.0044)   & (0.0061)   & (0.0043)   & (0.0049)   & (0.0050)   & (0.0054)   & (0.0028)   & (0.0034)    \\
High Subjective Quality Interaction                       & -0.0076    &            &            &            &            &            &            &            &            &             \\
                                       & (0.0066)   &            &            &            &            &            &            &            &            &             \\
High Subjective Quality                                     & -0.0879*** &            &            &            &            &            &            &            &            &             \\
                                       & (0.0098)   &            &            &            &            &            &            &            &            &             \\
Low Accuracy Interaction       &            & -0.0044    &            &            &            &            &            &            &            &             \\
                                       &            & (0.0081)   &            &            &            &            &            &            &            &             \\
Low Accuracy                     &            & -0.2045*** &            &            &            &            &            &            &            &             \\
                                       &            & (0.0120)   &            &            &            &            &            &            &            &             \\
Small Mask Interaction           &            &            & -0.0190**  &            &            &            &            &            &            &             \\
                                       &            &            & (0.0074)   &            &            &            &            &            &            &             \\
Small Mask                         &            &            & -0.0551*** &            &            &            &            &            &            &             \\
                                       &            &            & (0.0108)   &            &            &            &            &            &            &             \\
Low Entropy Interaction        &            &            &            & 0.0138*    &            &            &            &            &            &             \\
                                       &            &            &            & (0.0082)   &            &            &            &            &            &             \\
Low Entropy                      &            &            &            & 0.0264**   &            &            &            &            &            &             \\
                                       &            &            &            & (0.0120)   &            &            &            &            &            &             \\
1 Object Disappeared Interaction &            &            &            &            & -0.0016    &            &            &            &            &             \\
                                       &            &            &            &            & (0.0056)   &            &            &            &            &             \\
1 Object Disappeared               &            &            &            &            & 0.0086     &            &            &            &            &             \\
                                       &            &            &            &            & (0.0082)   &            &            &            &            &             \\
First Correct Interaction           &            &            &            &            &            & -0.0217*** &            &            &            &             \\
                                       &            &            &            &            &            & (0.0035)   &            &            &            &             \\
First Correct               &            &            &            &            &            & -0.0017    &            &            &            &             \\
                                       &            &            &            &            &            & (0.0037)   &            &            &            &             \\
Has Person Interaction              &            &            &            &            &            &            & 0.0130**   &            &            &             \\
                                       &            &            &            &            &            &            & (0.0060)   &            &            &             \\
Has Person                     &            &            &            &            &            &            & 0.0028     &            &            &             \\
                                       &            &            &            &            &            &            & (0.0088)   &            &            &             \\
Fast Completion Interaction          &            &            &            &            &            &            &            & -0.0108*   &            &             \\
                                       &            &            &            &            &            &            &            & (0.0066)   &            &             \\
Fast Completion                     &            &            &            &            &            &            &            & 0.0430***  &            &             \\
                                       &            &            &            &            &            &            &            & (0.0121)   &            &             \\
Mobile Interaction                   &            &            &            &            &            &            &            &            & 0.0264***  &             \\
                                       &            &            &            &            &            &            &            &            & (0.0067)   &             \\
Mobile                     &            &            &            &            &            &            &            &            & -0.0753*** &             \\
                                       &            &            &            &            &            &            &            &            & (0.0120)   &             \\
Right Placement Interaction           &            &            &            &            &            &            &            &            &            & 0.0088**    \\
                                       &            &            &            &            &            &            &            &            &            & (0.0043)    \\
Right Placement                  &            &            &            &            &            &            &            &            &            & -0.0108     \\
                                       &            &            &            &            &            &            &            &            &            & (0.0074)    \\
Constant                                & 0.8377***  & 0.8915***  & 0.8249***  & 0.7778***  & 0.8021***  & 0.8478***  & 0.8051***  & 0.7836***  & 0.8184***  & 0.8122***   \\
                                       & (0.0043)   & (0.0050)   & (0.0065)   & (0.0090)   & (0.0064)   & (0.0061)   & (0.0073)   & (0.0091)   & (0.0043)   & (0.0054)    \\
\hline
\\
\hline
$N$                                      & 51611      & 25637      & 25655      & 25868      & 51611      & 38454      & 51611      & 24963      & 51611      & 51611       \\
$R^2$                                     & 0.04       & 0.11       & 0.03       & 0.02       & 0.01       & 0.01       & 0.01       & 0.02       & 0.02       & 0.01        \\      \\

\hline
\end{tabular}
}
\caption{Ordinary least squares regression with image fixed effects evaluating image position on users' accuracy in identifying manipulated images. Robust standard errors clustered at the image level in parentheses. *, **, and *** indicates statistical significance at the 90, 95, and 99 percent confidence intervals, respectively. All columns drop users who submitted fewer than 10 guesses, drop all control images where nothing was removed, drop all guesses beyond each participants' 10th guess, and include image fixed effects. }
\label{table:ols_3}
\end{center}
\end{table}

\section*{Appendix II: Supplementary Information}

\begin{figure}[ht]
    \centering
    \includegraphics[width=0.75\textwidth]{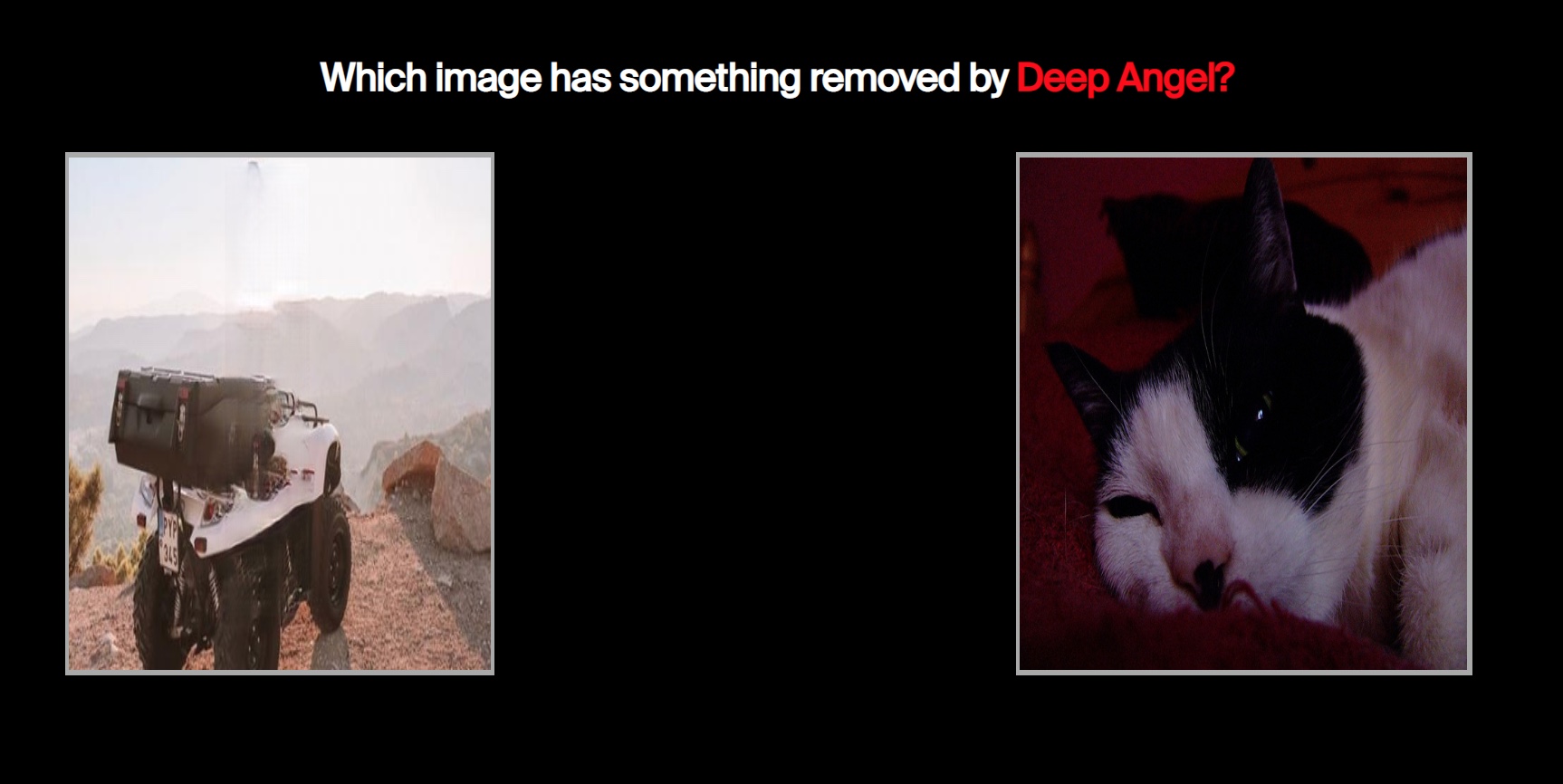}
    \caption{``Detect Fakes'' User Interfaces}
    \label{fig:uidetect}
\end{figure}

\begin{table}[h!]
\begin{center}
\begin{tabular}{lcccccc}
\textbf{Image Uploads} \\                         
                          \midrule
          \textbf{Object} & \textbf{Count} & \textbf{Order} \\
\midrule
\textbf{Person}             &      13450  &        1       \\
\textbf{Car}             &     1229  &        6       \\
\textbf{Dog}             &      1086  &        2       \\
\textbf{Cat}             &      1082  &        3       \\
\textbf{Elephant}             &      185  &        4       \\
\textbf{Bicycle}             &      158  &        7       \\
\textbf{Bird}             &      139  &        22       \\
\textbf{Tie}             &      120  &        31      \\
\textbf{Airplane}             &      106  &        13       \\
\textbf{Stop Sign}             &      99  &        8       \\
\bottomrule
\end{tabular}
\begin{tabular}{lcccccc}
                      \textbf{Instagram Directed Crawls} \\                         
                          \midrule              \textbf{Object} & \textbf{Count} & \textbf{Order} \\
\midrule
\textbf{Person}             &      6944  &        1       \\
\textbf{Cat}             &     725  &        2       \\
\textbf{Dog}             &      493  &        3       \\
\textbf{Elephant}             &      170 &        4       \\
\textbf{Car}             &     162  &        6       \\
\textbf{Bicycle}             &     71  &        7       \\
\textbf{Sheep}             &     52  &        5      \\
\textbf{Stop Sign}             &     31  &        8       \\
\textbf{Airplane}             &     29  &        13       \\
\textbf{Skateboard}             &     25  &        10       \\
\bottomrule
\end{tabular}

\end{center}
\caption{Top 10 \textit{Target Object Removal} Selections for Uploaded Images and Targeted Instagram Crawls on Deep Angel. Each selection of an Instagram username initiated a targeted crawl of Instagram for the three most recently uploaded images of the selected user.}
\label{table:selections}
\end{table}

\begin{figure}[h]
    \centering
\includegraphics[width=0.57\textwidth]{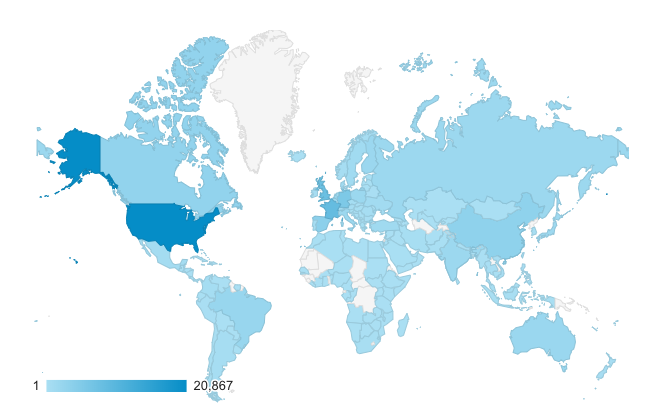}
    \caption{Heat map of the world showing how many users came from each country. 23\% of users are from the United States of America, 9\% from France, 9\% from United Kingdom, 6\% from Germany, 4\% from Spain, 3\% from China, 3\% from Canada, 2\% from Brazil, 2\% from Australia, and 2\% from Finland.}
    \label{fig:action32}
\end{figure}

\section*{Appendix IIIa: Detecting Manipulation by Image Features}

\begin{figure}[H]
    \centering
    
    \begin{subfigure}[t]{0.49\textwidth}
    \centering
    \includegraphics[width=0.99\textwidth]{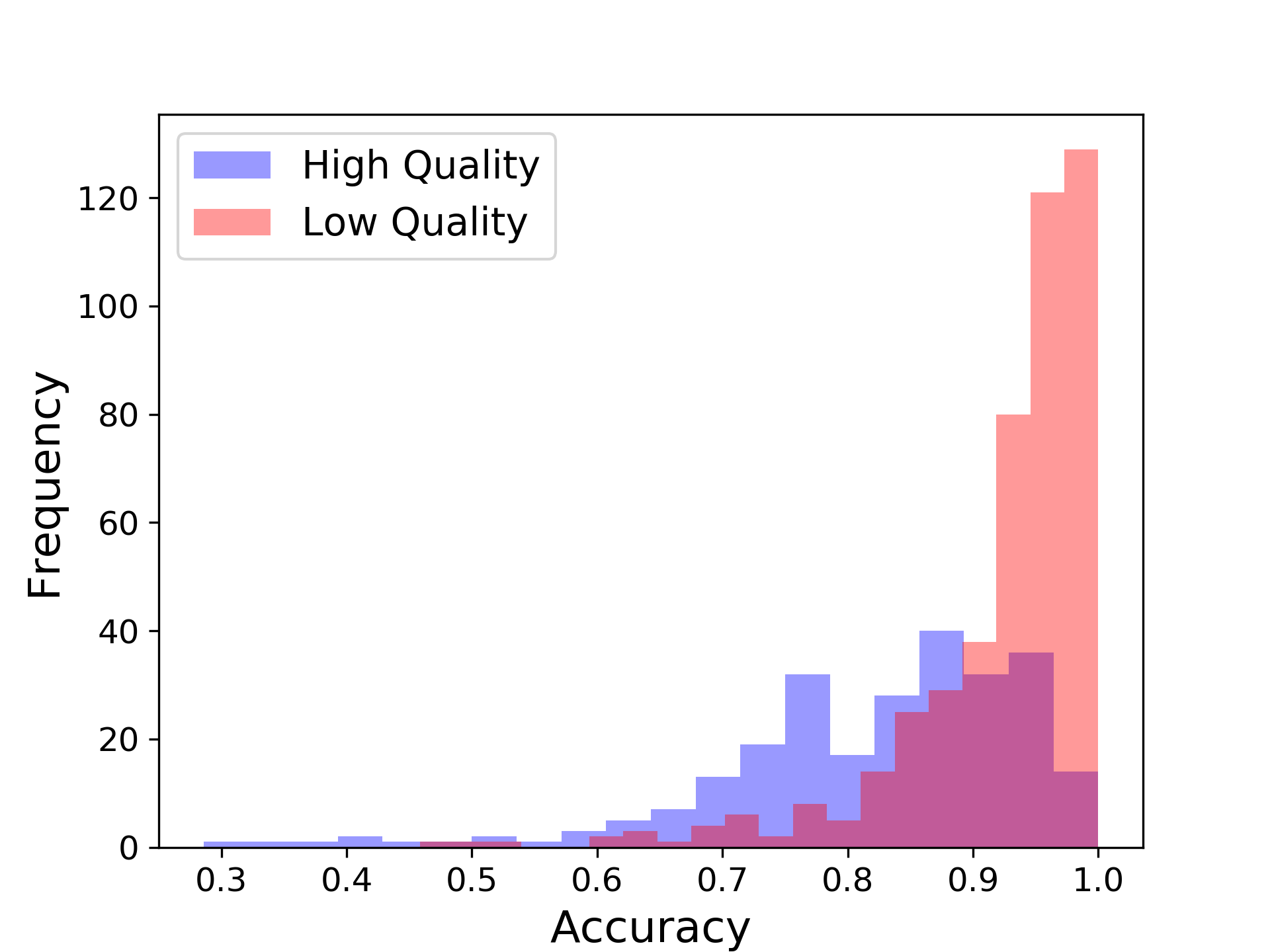}
    \caption{Subjective Quality}
    \end{subfigure}%
    ~
    \begin{subfigure}[t]{0.49\textwidth}
    \centering
    \includegraphics[width=0.99\textwidth]{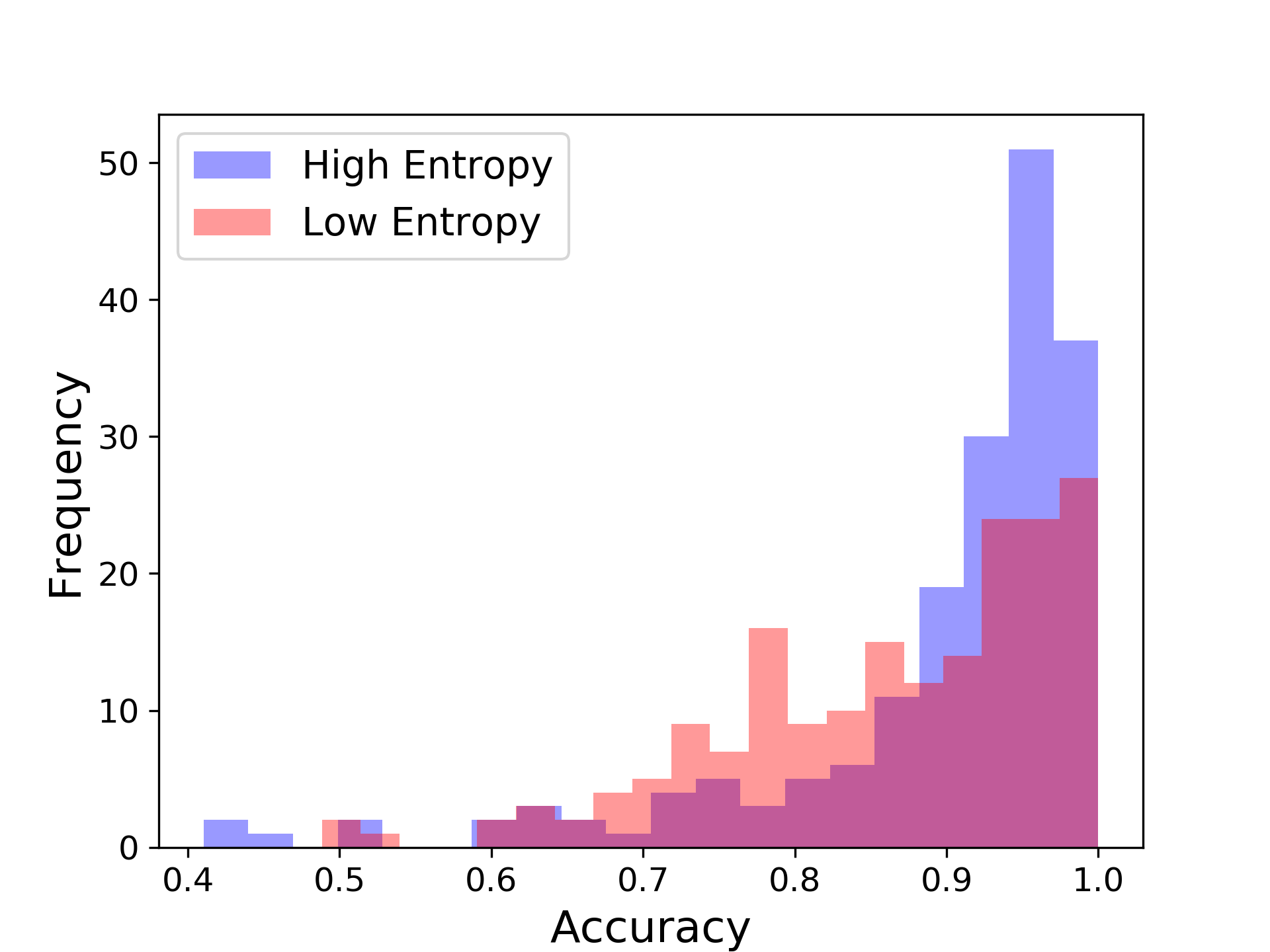}
    \caption{Entropy}
    \end{subfigure}%

    \begin{subfigure}[t]{0.49\textwidth}
    \centering
    \includegraphics[width=0.99\textwidth]{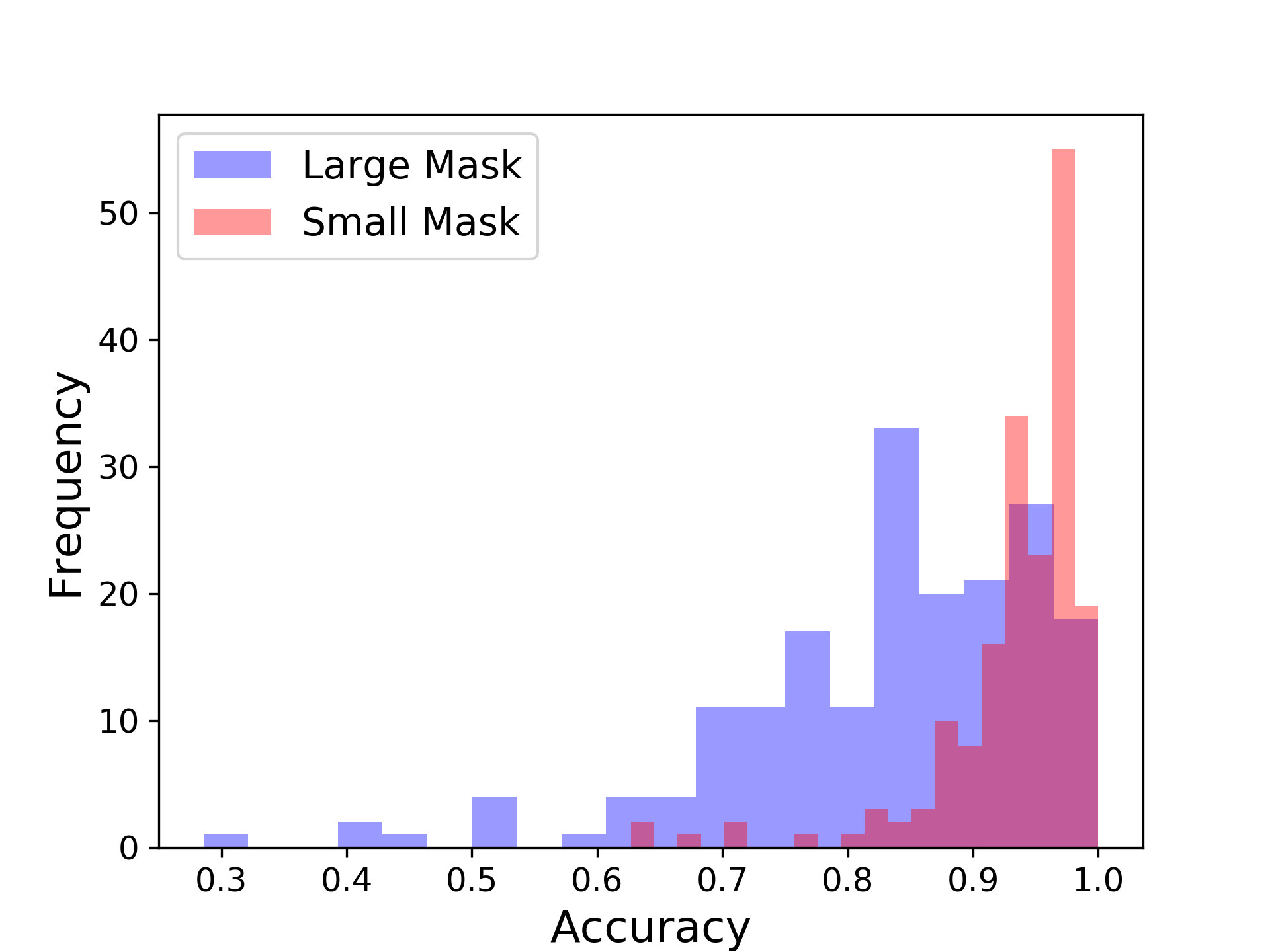}
    \caption{Area Transformed}
    \end{subfigure}%
    ~
    \begin{subfigure}[t]{0.49\textwidth}
    \centering
    \includegraphics[width=0.99\textwidth]{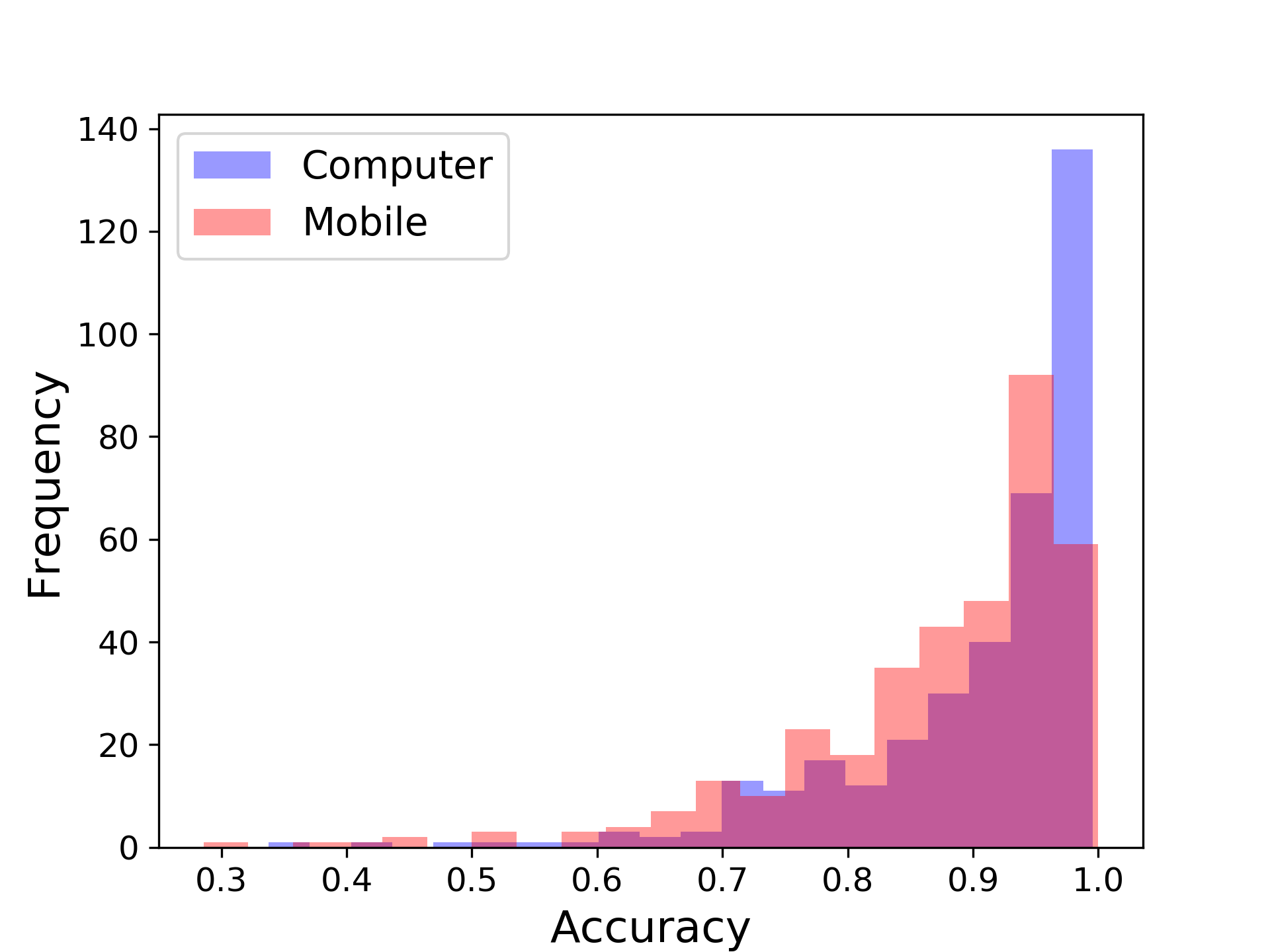}
    \caption{Device}
    \end{subfigure}%
    \caption{Histograms comparing the distribution of mean accuracy scores across images for four image features: subjective quality, entropy, image mask size, and user device.   }
    \label{fig:histowow}
\end{figure}

Figure~\ref{fig:histowow} shows the distribution of mean accuracy scores across images varies by image features. For example, the set of images rated as high quality have a mean accuracy score of 82\% whereas the images rated as low quality have a mean accuracy score of 92\%. Likewise, the high quality images have a wider variance in accuracy scores than the low quality images. Another example indicating image characteristics' role in image detection is the differential distribution in accuracy across images with high and low entropy and large and small masks. The distributions in mean accuracy scores across devices is a less stark contrast than the other distribution comparisons, but here, we see that images viewed on mobile devices perform are slightly less accurately identified.  

\section*{Appendix IIIb: Unanchored Object Conjuring}
While posing a risk to information online, these generative AI systems can also offer new possibilities for creative expression. For example, the Creative Adversarial Network learns art by its styles and generates new art by deviating from the styles' norms \cite{DBLP:journals/corr/ElgammalLEM17}. Likewise, interactive GANs (iGANS) can augment human creativity for artistic and design applications \cite{carter2017using}.

Thus, if objects can be plausibly removed from images, then it is reasonable to imagine objects can be plausibly generated in an image from which they never existed. As an extension to the Deep Angel pipeline, we approached adding objects to images using image-to-image translation with conditional adversarial networks \cite{DBLP:journals/corr/IsolaZZE16}. Since these neural networks learn a mapping from an input to an output image, we can train an image-to-image model using the manipulated images as inputs and the original submissions are outputs. While the model does not re-appear objects as they were, the model produces resemblances of the missing objects. Images produced by Deep Angel and AI Spirits (the reversal of Deep Angel) are on display at the online art gallery hosted by the 2018 NeurIPS Workshop on Machine Learning for Creativity and Design. As large-scale, paired datasets of creative content (such as the one presented here) become increasingly common, and neural network architectures for content generation become more powerful, automated object insertion into existing media will become a rich area for future work.

\subsection*{Model}

With image-to-image translation, a latent representation of the structure an image can be efficiently expressed in and generated for different contexts \cite{carter2017using, DBLP:journals/corr/IsolaZZE16,wang2018high}. This latent structure is encode in information like edges, shape, size, texture, and color that are anchored across contexts. By applying image-to-image translation to the results of the \textit{Target Object Removal} pipeline, we force the model to learn both the structural representation for removed objects and their contextual location. We call this process unanchored object conjuring. 

For the unanchored object conjuring extension, the global component ($G_1$) ($7 \times 7$ Convolution-InstanceNorm ReLU layer with 32 filters and stride 1 \cite{ulyanov2017instance}) in the top right of Figure~\ref{arch} is first trained on downsampled images, then local component ($G_2$) is concatenated to $G_1$ and they are jointly trained on full resolution images. We follow the original pix2pixHD loss function which takes the form 
\begin{align*}
\min_G \left(\left(\max_{D_1, D_2, D_3} \sum_{k=1,2,3} \mathcal{L}_{GAN}(G,D_k)\right) + \lambda_{VGG}\mathcal{L}_{VGG}(G(x),y)\right) + \lambda_{fm}\sum_{k=1,2,3} \mathcal{L}_{fm}(G,D_k)
\end{align*}
where $\mathcal{L}_{GAN}\left(\cdot\right)$ is adversarial loss \cite{isola2017image}, $\mathcal{L}_{fm}\left(\cdot\right)$ is the feature matching loss pix2pixHD used to stabilize training and $\mathcal{L}_{VGG}\left(\cdot\right)$ is the perceptual loss based on VGG features \cite{johnson2016perceptual, simonyan2014very}. We train the model using the Adam solver with a learning rate $\eta =  0.0002$ for 200 epochs \cite{kingma2014adam}. $\eta$ is fixed for the first half of training (epochs 0 to 100) and then $\eta$ linearly decays to 0 for the second half (epochs 101 to 200). All weights were initialized by sampling from a Gaussian distribution with $\mu =  0$ and $\sigma =  0.02$  \cite{wang2018high}. We used a PyTorch implementation with a batch size of 4 on an Nvidia Geforce GTX Titan X with 8 cores.

\subsection*{Data}
\begin{figure}[h]
  \centering
 \includegraphics[width=0.67\textwidth]{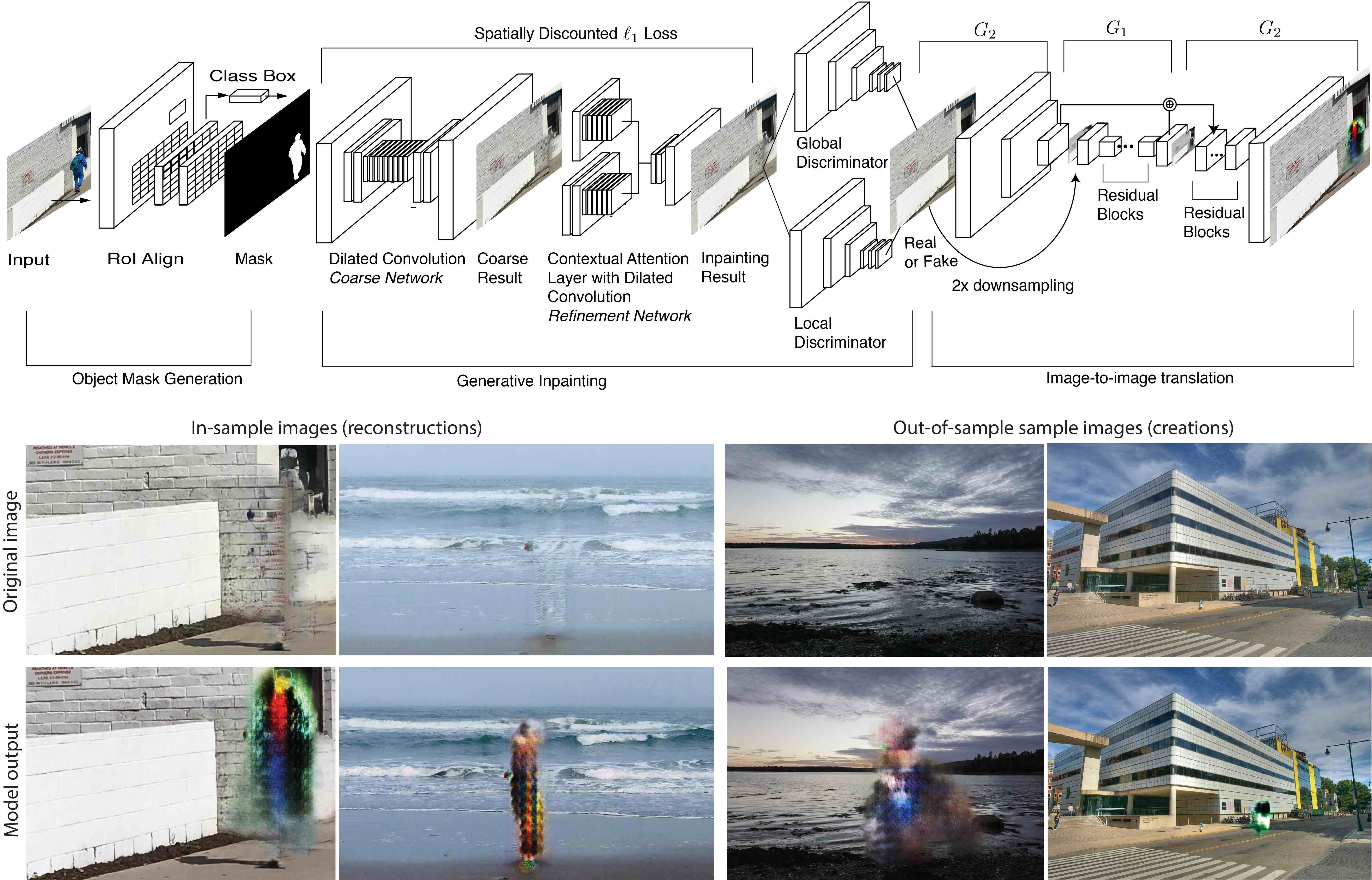}
  \caption{The top row displays 4 input images and the bottom rows displays the modeled output based on the unanchored object conjuring pipeline. The images on the left are considered reconstructions because they are part of the paired training sample and the images on the right are considered creations because they are not part of the training dataset.}
  \label{arch}
\end{figure}

We filtered all images uploaded to Deep Angel to 5,634 images where people were selected to be removed. We manually filtered these images to the 1000 best manipulations based on qualitative judgements. Then, we resized and cropped images to $1024 \times 1024$. We trained these images following the pix2pixHD image-to-image translation architecture, which yields improved photorealism due to its coarse-to-fine generators, multi-scale discrimination and improved adversarial loss~\cite{wang2018high}. Figure~\ref{arch} shows the architecture for this extended pipeline.

This unanchored object conjuring technique can be used to create a new class of art that combine existing photographs with GAN-style imagery. In addition, the reconstructions provide a technique for interpreting the model and the underlying dataset by revealing where removed objects systematically appear. 

\end{document}